\documentclass[11pt]{article}

\usepackage[preprint]{acl}

\usepackage{times}
\usepackage{latexsym}
\usepackage[T1]{fontenc}
\usepackage[utf8]{inputenc}
\usepackage{microtype}

\usepackage{graphicx}
\usepackage{booktabs}
\usepackage{multirow}
\let\PlainTabular\tabular
\let\PlainEndTabular\endtabular
\usepackage{colortbl}
\usepackage{arydshln}
\let\ColorTabular\tabular
\let\ColorEndTabular\endtabular
\let\OrigMaketitle\maketitle
\renewcommand{\maketitle}{%
  \let\tabular\PlainTabular
  \let\endtabular\PlainEndTabular
  \OrigMaketitle
  \let\tabular\ColorTabular
  \let\endtabular\ColorEndTabular
}

\usepackage{amsmath}
\usepackage{amssymb}
\usepackage{bm}

\graphicspath{{images/}}

\definecolor{HeaderRow}{RGB}{230,238,247}   
\definecolor{EmphRow}{RGB}{250,236,217}     

\newcommand{\Ru}{R_{u}}
\newcommand{\Rs}{R_{s}}
\newcommand{\Aneedle}{A_{\mathrm{needle}}}
\newcommand{\Ahaystack}{A_{\mathrm{haystack}}}
\newcommand{\Acontext}{A_{\mathrm{context}}}
\newcommand{\Hsafety}{\mathcal{H}_{\mathrm{safety}}}
\newcommand{\Funsafe}{F_{1}^{\mathrm{unsafe}}}
\newcommand{\qstar}{q^{*}}

\title{LongGuard: Mechanistic Analysis and Training-Free Mitigation of Long-Context Failure in Safety Guardrails}

\author{Ziyang Chen\textsuperscript{1,2}, Xing Wu\textsuperscript{1,2}, and Songlin Hu\textsuperscript{1,2}\thanks{Corresponding author.} \\
  \textsuperscript{1} Institute of Information Engineering, Chinese Academy of Sciences, Beijing, China \\
  \textsuperscript{2} School of Cyber Security, University of Chinese Academy of Sciences, Beijing, China \\
  \texttt{\{chenziyang,wuxing,husonglin\}@iie.ac.cn} \\}

\begin{document}
\maketitle

\begin{abstract}
Safety guardrails serve as the last line of defense against harmful inputs and outputs of large language models (LLMs), yet they are trained and evaluated almost exclusively on short text. We present \textbf{LongGuard}, a framework that evaluates, mechanistically analyzes, and mitigates long-context guardrail failure. We formulate the task as \emph{Safety Needle-in-a-Haystack} (SafetyNIAH) over a 0.25k--32k length grid; across 15 mainstream guardrails, unsafe recall drops monotonically by more than 50\% on average, and a paired Benign-Fill vs.\ Needle-Repeat design attributes the failure to proportional dilution of the unsafe needle rather than to absolute length. A three-layer attention--logit--behavior analysis on six guardrails locates the mechanism: attention mass on the unsafe needle is diluted, the unsafe-over-safe logit margin is compressed in lockstep, and the detection decision collapses accordingly, with this attention$\to$logit$\to$behavior chain remaining consistent after partialling out length. We further isolate a sparse set of \emph{guard-specialized retrieval heads} that exhibit \emph{partial specificity} relative to their base models. Building on the analysis, we propose two training-free mitigations -- \emph{Chunked Detection} (CD) and \emph{Attention-Head Sharpening} (AHS) -- and a deployment protocol, \emph{Context-Aware Hyperparameter Routing} (CAHR), that selects configurations by context length and audit side. Across five benchmarks spanning synthetic data, long-context attacks, and reasoning-model outputs, CAHR-CD and CAHR-AHS improve the six-guardrail average by 22\% and 13\%, respectively. Code and data are available online.\footnote{\url{https://github.com/caskcsg/LongGuard}}
\end{abstract}

\begin{figure}[t]
  \centering
  \includegraphics[width=\columnwidth]{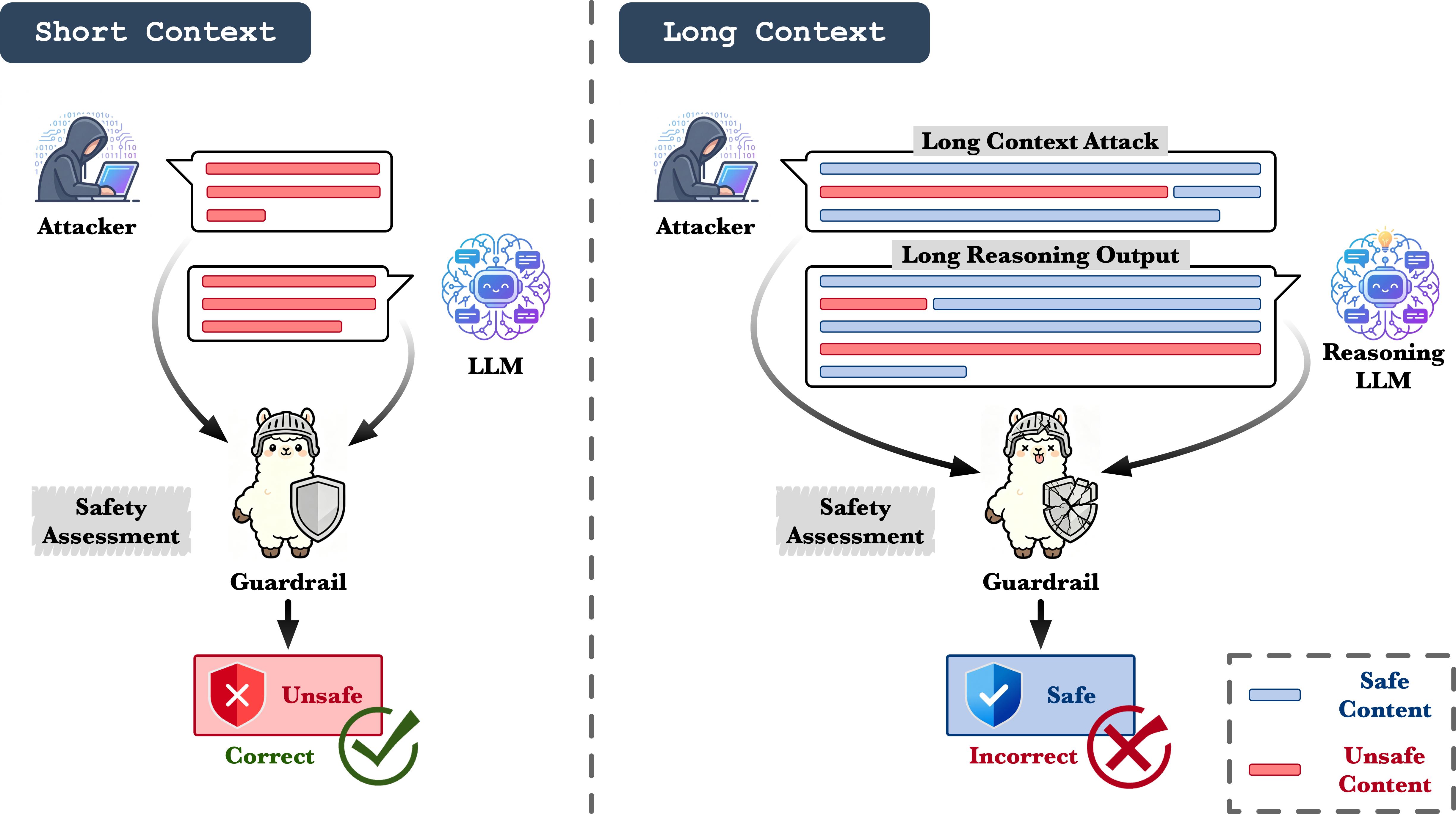}
  \caption{Long-context attacks (on the input side) and long reasoning outputs (on the output side) pose new challenges for safety guardrails trained on short text.}
  \label{fig:overview}
\end{figure}

\section{Introduction}

Safety guardrails serve as real-time moderators of LLM inputs and outputs and have become indispensable for LLM safety, with multiple generations~\cite{inan2023llama,han2024wildguard,zeng2024shieldgemma,kumar2025polyguard,zhao2025qwen3guard,lin2026yufeng} widely deployed in short-text scenarios. The sequences they moderate, however, are growing longer (Figure~\ref{fig:overview}): Many-shot Jailbreaking~\cite{anil2024many} and NINJA Attack~\cite{shah2025jailbreaking} distribute harmful instructions across long prompts to bypass safety alignment, and reasoning models such as DeepSeek-R1~\cite{guo2025deepseek} and Qwen3~\cite{yang2025qwen3} produce thousands of intermediate tokens that may carry latent risks. Together, they shift the deployment distribution of guardrails from short text to long context. While long-context safety of LLMs themselves has begun to receive attention~\cite{lu2025longsafety,huang2025longsafety}, whether guardrails still play the role of the last line of defense under long context remains open.

\begin{figure*}[t!]
  \centering
  \includegraphics[width=\textwidth]{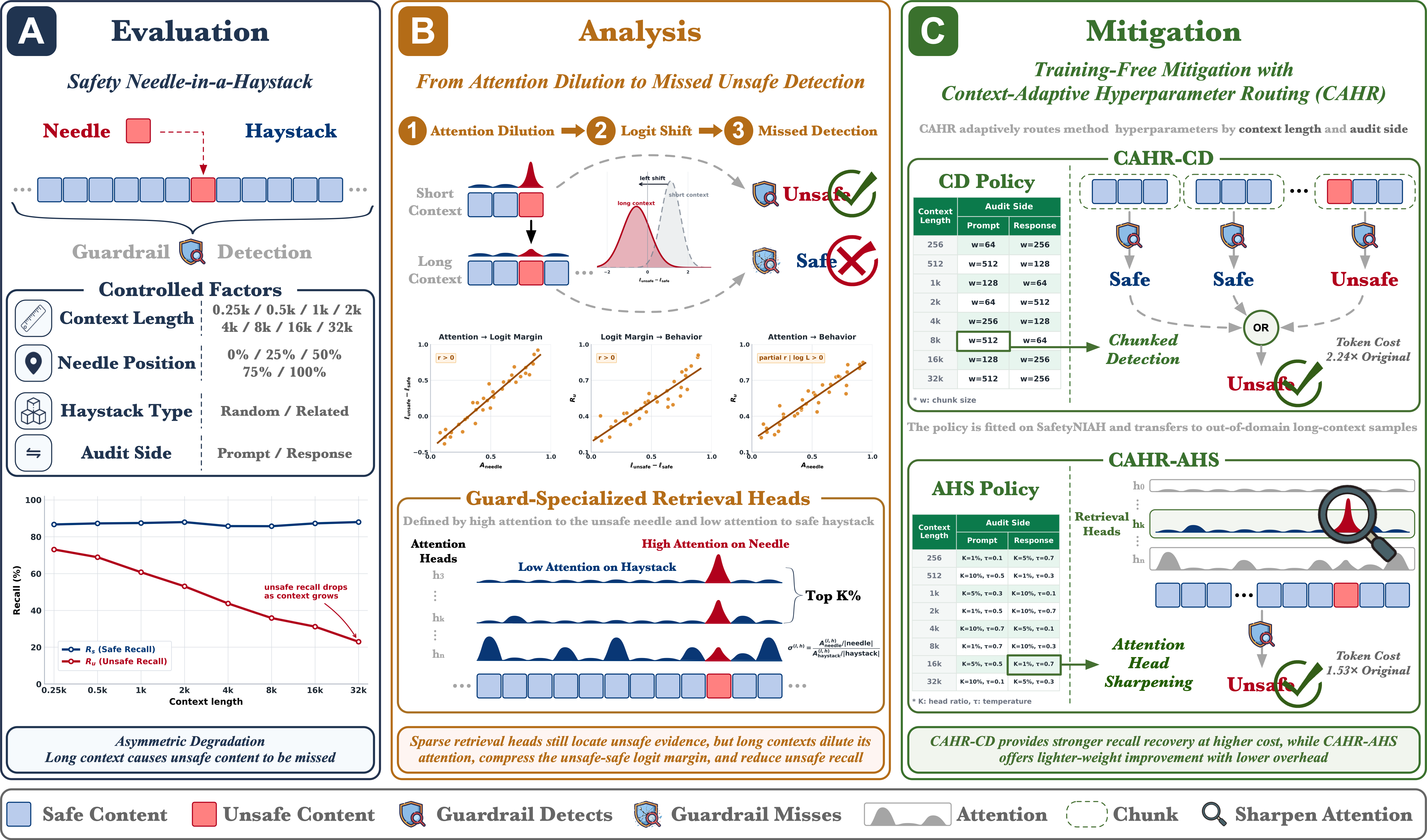}
  \caption{Overview of LongGuard: evaluation, mechanistic analysis, and training-free mitigation.}
  \label{fig:framework}
\end{figure*}

We organize the paper around three research questions. \textbf{(RQ1)} Does guardrail detection degrade systematically with context length? \textbf{(RQ2)} If so, what is the internal mechanism? \textbf{(RQ3)} Can the mechanism guide \emph{training-free} mitigations? We answer them in \textbf{LongGuard}, a unified evaluation, analysis, and mitigation framework with three contributions (Figure~\ref{fig:framework}).

\textbf{Evaluation.} We formulate the task as Safety Needle-in-a-Haystack (SafetyNIAH) and synthesize a benchmark over a 0.25k--32k length grid with controlled needle type, position, and haystack type; a paired Benign-Fill / Needle-Repeat design separates proportional dilution from absolute length. Across 15 guardrails, unsafe recall declines monotonically, while safe recall remains flat—an asymmetric degradation driven mainly by dilution.

\textbf{Analysis.} On six guardrails, a three-layer attention--logit--behavior framework localizes the failure to dilution of unsafe evidence by the haystack. It isolates a sparse set of \emph{guard-specialized retrieval heads} that exhibit \emph{partial specificity} relative to their base models.

\textbf{Mitigation.} Two training-free methods follow: \emph{Chunked Detection} (CD) chunks the input to raise local evidence density, and \emph{Attention-Head Sharpening} (AHS) sharpens the softmax temperature on the retrieval heads; \emph{Context-Aware Hyperparameter Routing} (CAHR) selects hyperparameters by context length and audit side. Across five benchmarks, they restore long-context detection and form a performance--cost trade-off.

\section{Related Work}
\label{sec:related}

\paragraph{Long-context safety of LLMs.}
Prior work studies how long context affects the safety behavior of LLMs themselves: Many-shot Jailbreaking~\cite{anil2024many} and NINJA Attack~\cite{shah2025jailbreaking} attack alignment with in-context demonstrations or needle-in-a-haystack patterns, and LongSafety~\cite{lu2025longsafety,huang2025longsafety} evaluates long-context safe generation. We instead study the complementary deployment-side question -- whether guardrails still reliably block unsafe content under long prompts and reasoning outputs.

\paragraph{Safety guardrails and benchmarks.}
Open-source guardrails have evolved along data, taxonomy, and modality axes -- spanning fine-tuned harmful-content classifiers, reasoning-chain guards, multilingual variants, and configurable strict/loose modes -- and the safety benchmarks evaluating them now cover jailbreak, harmful instructions, toxicity, refusal, and output risks (full lists in Appendices~\ref{app:models} and~\ref{app:data}). Both substantially improve detection in standard settings, yet they remain almost exclusively limited to short text; LongGuard targets this shared blind spot, and SafetyNIAH reuses their safety semantics while injecting controlled length, position, and haystack dimensions.

\paragraph{Attention and safety mechanisms.}
Recent mechanistic studies identify sparse attention heads tied to safety behavior and probe them via entropy or realignment~\cite{zhou2025role, ostmeier2026attention}, but mainly on short inputs of general LLMs. We extend this perspective to long-input guardrail detection and use base--guard comparison to localize the shaping effect of guard fine-tuning.

\section{SafetyNIAH: Task and Benchmark}
\label{sec:safetyniah}

This section defines the task and the construction of long-context guardrail samples used in RQ1.

\subsection{Task Formulation}

We formulate long-context harmful content detection as the \textbf{Safety Needle-in-a-Haystack (SafetyNIAH)} task: a short, labeled sample (needle) is embedded in a neutral long context (haystack); the guardrail must keep the original decision from being diluted. Let the guardrail be
\begin{equation}
f_{\theta}: \mathcal{X} \rightarrow \{\mathrm{safe}, \mathrm{unsafe}\},
\label{eq:guardrail}
\end{equation}
where $\mathcal{X}$ is the text space to moderate: at the input side, $x$ is a prompt, and at the output side, $x=(q, r)$ is the concatenation of the user query and the model response. Given a needle pool $\mathcal{N}=\{(x_n, y_n, t_n)\}$ with labels $y_n\in\{\mathrm{safe},\mathrm{unsafe}\}$ and audit sides $t_n\in\{\mathrm{prompt},\mathrm{response}\}$, SafetyNIAH controls three variables: context length $L$, needle position $p$, and haystack type $\rho$, and synthesizes
\begin{equation}
x_{\mathrm{final}} = \mathcal{C}(x_n;\, L, p, \rho, t_n),
\label{eq:safetyniah}
\end{equation}
checking whether $f_{\theta}(x_{\mathrm{final}})$ still agrees with $y_n$.

The construction decomposes long-context degradation into two external factors: \emph{proportional dilution} (the share of needle tokens decreases) and \emph{absolute-length effect} (instability induced by length itself). To separate them, we build two paired sets. The \textbf{Benign-Fill} main benchmark surrounds a needle with a neutral haystack so that dilution and length grow together. The \textbf{Needle-Repeat} negative control replicates an unsafe needle to the target length, removing dilution and isolating the length.

\subsection{Needle Pool}

We collect samples from 17 guardrail benchmarks (Appendix~\ref{app:data}). We then stratify by target length $\mathcal{L}=\{256, 512, 1{,}024, 2{,}048, 4{,}096, 8{,}192, \allowbreak 16{,}384, 32{,}768\}$ words, label, and audit side, sampling up to 50 instances per cell. The resulting needle pool contains around 15.2k samples.

\subsection{Neutral Haystack Pool}

To prevent the haystack from carrying safety risks that contaminate the needle label, we build a neutral haystack pool as follows. We uniformly sample 500k articles from English Wikipedia,\footnote{\url{https://huggingface.co/datasets/wikimedia/wikipedia/viewer/20231101.en}} concatenate consecutive sentences into chunks of at most 2{,}048 characters, and apply strict filtering with a set $\mathcal{G}_{\mathrm{filter}}$ of guardrails (Appendix~\ref{app:models}): every chunk is scored independently under both prompt and response settings, and a chunk is dropped if \emph{any} guardrail under \emph{any} setting marks it unsafe. The cleaned pool is denoted $\mathcal{D}_{\mathrm{clean}}$.

\subsection{Long-Context Sample Synthesis}

\paragraph{Benign-Fill main benchmark.} Given a needle $x_n$, context length $L\in\mathcal{L}$, needle position $p\in\{0\%, 25\%, 50\%, 75\%, 100\%\}$, and haystack type $\rho\in\{\mathrm{Random},\mathrm{Related}\}$, we form
\begin{equation}
x_{\mathrm{final}} = H_{\mathrm{before}}\oplus x_n \oplus H_{\mathrm{after}},
\label{eq:benign-fill}
\end{equation}
where the word budgets are $w_{\mathrm{before}}=\lfloor (L-|x_n|)\,p\rfloor$ and $w_{\mathrm{after}}=L-|x_n|-w_{\mathrm{before}}$. Both $H_{\mathrm{before}}$ and $H_{\mathrm{after}}$ are produced by concatenating chunks from $\mathcal{D}_{\mathrm{clean}}$. Random and Related share the same $(x_n, L, p)$ and form strict pairs: Random samples chunks uniformly without replacement from $\mathcal{D}_{\mathrm{clean}}$ with a needle-specific seed; Related encodes needle and candidate chunks with \texttt{jina-embeddings-v3}~\cite{sturua2024jina} and retrieves the most similar chunks as the haystack. The final Benign-Fill set has 30{,}400 samples; the full distribution is summarized in Appendix~\ref{app:data}.

\paragraph{Needle-Repeat negative control.} Benign-Fill couples dilution with length. To isolate length, we replicate each unsafe needle to the target length, so that the full context consists of copies of the needle. The self-fill zeroes out dilution while keeping length, and the needle position/haystack-type axes are no longer applicable. Needle-Repeat uses only unsafe needles, yielding 8{,}800 samples.

\section{Long-Context Degradation and Causal Decomposition}
\label{sec:degradation}

This section answers RQ1. Section~\ref{sec:degradation-main} reports an asymmetric degradation of unsafe recall across 15 guardrails. Section~\ref{sec:degradation-cause} decomposes the external cause into proportional dilution and an absolute-length residual through Benign-Fill / Needle-Repeat. Additional ablations are deferred to Appendix~\ref{app:behavior-ablation}.

\begin{table*}[t!]
    \centering
  \resizebox{\textwidth}{!}{%
  \begin{tabular}{lcccccccccc>{\columncolor{gray!12}}ccc}
        \toprule
    \rowcolor{white}
    \multirow{2}{*}{Model}
      & \multicolumn{2}{c}{0.25k} & \multicolumn{2}{c}{1k} & \multicolumn{2}{c}{4k} & \multicolumn{2}{c}{16k} & \multicolumn{2}{c}{32k} & \multirow{2}{*}{$\Delta\Ru$} & \multirow{2}{*}{$\Delta\Rs$} & \multirow{2}{*}{\texttt{inv@32k}} \\
    \cmidrule(lr){2-3}\cmidrule(lr){4-5}\cmidrule(lr){6-7}\cmidrule(lr){8-9}\cmidrule(lr){10-11}
    \rowcolor{white}
      & $\Ru$ & $\Rs$ & $\Ru$ & $\Rs$ & $\Ru$ & $\Rs$ & $\Ru$ & $\Rs$ & $\Ru$ & $\Rs$ & & & \\
        \midrule
    WildGuard-7B & 86.36 & 83.75 & 76.32 & 87.25 & 13.73 & 66.00 & 2.27 & 58.94 & 0.09 & 56.94 & $-$86.27 & $-$26.81 & 39.71\% \\
    GPT-OSS-SafeGuard-20B & 83.73 & 81.12 & 81.59 & 79.56 & 81.27 & 80.38 & 78.59 & 79.44 & 68.91 & 79.94 & $-$14.82 & $-$1.19 & 6.50\% \\
    NemotronGuardV2-8B & 64.41 & 92.12 & 47.27 & 87.88 & 17.09 & 81.19 & 21.68 & 86.12 & 25.09 & 85.00 & $-$39.32 & $-$7.13 & 7.45\% \\
    NemotronGuardV3-8B & 78.05 & 86.31 & 69.14 & 74.00 & 36.91 & 61.06 & 14.09 & 54.87 & 16.86 & 54.06 & $-$61.18 & $-$32.25 & 36.03\% \\
    NemotronReasoning-4B & 77.23 & 89.25 & 59.77 & 94.50 & 23.55 & 97.69 & 19.77 & 98.56 & 14.55 & 98.94 & $-$62.68 & $+$9.69 & 0.03\% \\
    LlamaGuard3-8B & 66.14 & 92.12 & 37.73 & 95.81 & 25.86 & 97.31 & 15.68 & 98.06 & 14.50 & 98.25 & $-$51.64 & $+$6.13 & 0.00\% \\
    LlamaGuard4-12B & 55.50 & 92.62 & 25.05 & 97.94 & 9.09 & 99.00 & 6.95 & 99.06 & 7.27 & 98.62 & $-$48.23 & $+$6.00 & 0.29\% \\
    ShieldGemma-9B & 41.45 & 95.75 & 21.55 & 98.44 & 6.73 & 99.19 & 0.00 & 100.00 & 0.00 & 100.00 & $-$41.45 & $+$4.25 & 0.00\% \\
    PolyGuard-7B & 91.86 & 67.19 & 89.23 & 69.75 & 86.95 & 71.12 & 74.41 & 73.38 & 50.32 & 77.81 & $-$41.55 & $+$10.62 & 0.00\% \\
    GuardReasoner-8B & 88.36 & 81.50 & 78.45 & 83.75 & 63.32 & 86.25 & 50.18 & 89.19 & 44.50 & 89.12 & $-$43.86 & $+$7.63 & 0.29\% \\
    Qwen3Guard-Gen-8B (strict) & 87.86 & 84.75 & 84.55 & 84.00 & 79.95 & 86.31 & 62.82 & 91.31 & 33.45 & 95.25 & $-$54.41 & $+$10.50 & 0.00\% \\
    Qwen3Guard-Gen-8B (loose) & 56.50 & 96.38 & 44.45 & 96.44 & 39.27 & 96.00 & 31.14 & 96.56 & 10.32 & 98.94 & $-$46.18 & $+$2.56 & 0.00\% \\
    Qwen3Guard-Stream-8B (strict) & 84.59 & 78.50 & 79.82 & 80.44 & 68.68 & 83.00 & 27.55 & 93.00 & 18.36 & 95.19 & $-$66.23 & $+$16.69 & 0.00\% \\
    Qwen3Guard-Stream-8B (loose) & 51.45 & 89.44 & 38.95 & 91.44 & 34.41 & 90.94 & 12.45 & 96.19 & 8.09 & 97.12 & $-$43.36 & $+$7.69 & 0.00\% \\
    YuFeng-XGuard-8B & 83.00 & 89.44 & 77.50 & 90.69 & 70.09 & 91.25 & 50.14 & 94.31 & 32.41 & 94.56 & $-$50.59 & $+$5.13 & 0.18\% \\
        \bottomrule
    \end{tabular}}
  \caption{Unsafe/Safe recall of 15 guardrails on Benign-Fill at five representative context lengths. The $\Delta$ columns report 32k$-$0.25k; \texttt{inv@32k} is the fraction of unparseable outputs at 32k mapped to the opposite class. Three-class models with a \emph{controversial} tier are split into \emph{-strict}/\emph{-loose} variants.}
  \label{tab:main-recall}
\end{table*}

\subsection{Asymmetric Degradation of Unsafe Recall}
\label{sec:degradation-main}

Across the 15 rows of Table~\ref{tab:main-recall}, $\Ru$ decreases monotonically from 0.25k to 32k with a cross-model mean $\Delta\Ru\!=\!-50.12\%$, while $\Delta\Rs$ averages $-1.30\%$. Long context primarily weakens the detection of unsafe content rather than producing a uniform drop in both classes. The negative $\Rs$ outliers come from parsing failures: four models have 6.5\%--39.7\% \texttt{inv@32k} outputs, which are counted as false alarms on safe ground truth. Long-context guardrail failures manifest in two patterns: unsafe samples being judged safe, and certain models failing to produce protocol-conforming outputs.

\subsection{Proportional Dilution vs.\ Absolute Length}
\label{sec:degradation-cause}

Table~\ref{tab:main-recall} confirms the degradation but does not separate proportional dilution from absolute-length effects. We use the Benign-Fill / Needle-Repeat pair as a decomposition.

\begin{figure}[t]
  \centering
  \includegraphics[width=\columnwidth]{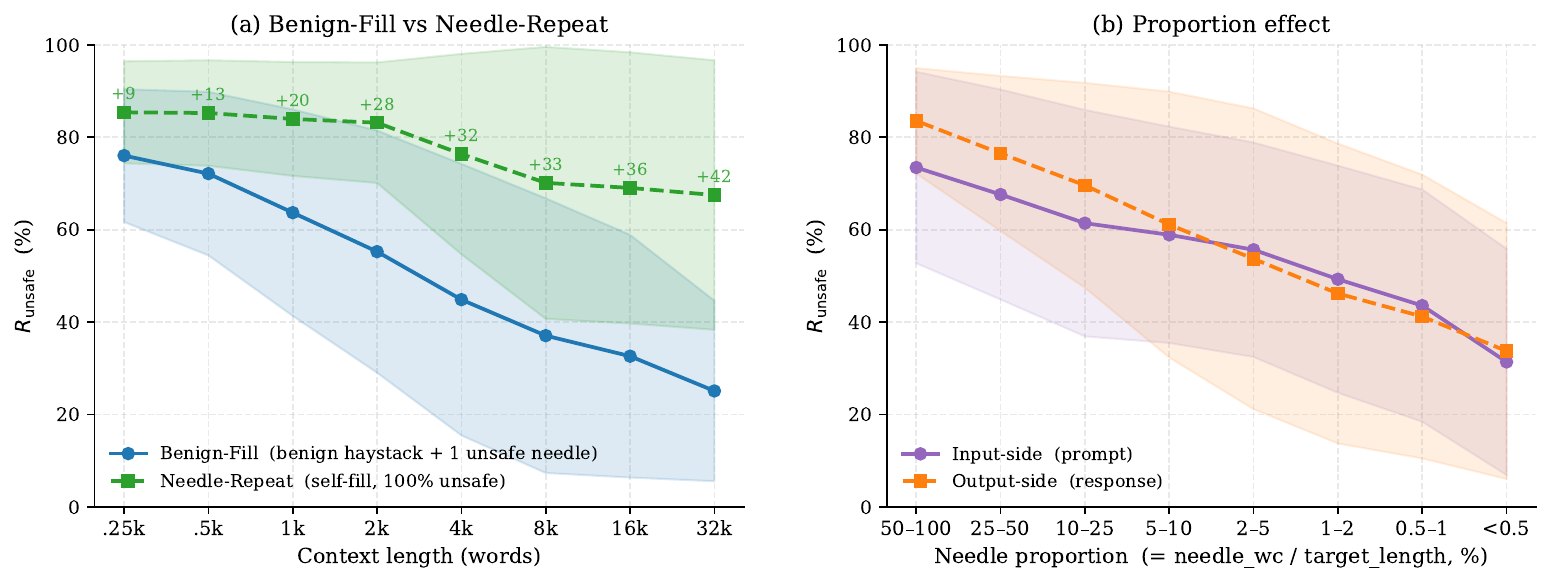}
  \caption{Two external causes of long-context degradation. (a) Benign-Fill vs.\ Needle-Repeat. (b) Relationship between needle share $s\!=\!|x_n|/L$ and $\Ru$ in Benign-Fill.}
  \label{fig:benign-vs-needle}
\end{figure}

\paragraph{Dilution is the dominant cause.} As Figure~\ref{fig:benign-vs-needle}(a) shows, Benign-Fill $\Ru$ falls from 76.04\% at 0.25k to 25.10\% at 32k ($\Delta=-50.94\%$), while Needle-Repeat falls from 85.41\% to 67.50\% ($\Delta=-17.91\%$). The Needle-Repeat\,$-$\, Benign-Fill gap widens from $+$9.4\% to $+$42.4\%, indicating that natural long-context degradation is mainly driven by dilution; the residual drop of Needle-Repeat reflects length-related effects (e.g.\ context-window limits, positional extrapolation).

\paragraph{Needle share monotonically tracks $\Ru$.} Define the needle share as $s\!=\!|x_n|/L$, i.e.\ the fraction of words taken up by the unsafe needle in the final $L$-word input. Figure~\ref{fig:benign-vs-needle}(b) shows that lower $s$ predicts lower $\Ru$: as $s$ decreases from $[50\%,100\%]$ to $[0,0.5\%)$, prompt-side $\Ru$ drops from 73.7\% to 31.7\% and response-side from 83.5\% to 33.6\%. This further supports dilution as the primary cause.

Dilution causes a substantially larger loss than absolute length and is concentrated on unsafe samples. The remainder of the paper therefore focuses on dilution-induced failure and centers its analysis on the unsafe class.

\section{Mechanism of Dilution-Driven Failure}
\label{sec:mechanism}

This section answers RQ2. Section~\ref{sec:degradation} attributes the degradation to dilution at the behavior level. We now analyze its internal mechanism on six representative full-attention guardrails. All metrics are computed on unsafe samples and weighted by sample counts when merging prompt and response audit sides.

\subsection{From Attention to Logit to Behavior}

\paragraph{The decision token.} We use $\qstar$ for the decision token of each sample, defined as the \emph{last non-padding query token} in the input rendered by the native chat template of each guardrail; its position is the index at which the model produces the classification logits in the forward pass.

\subsubsection{Attention: needle attention is diluted}

We decompose the attention of $\qstar$ over the input sequence $T$ into three spans: \emph{needle} (annotated unsafe needle tokens), \emph{haystack} (distractor content around the needle), and \emph{context}~$=T\setminus(\text{needle}\cup\text{haystack})$ (e.g., BOS, system prompt, chat template). For each layer and head, we take the attention weights from $\qstar$ to all tokens, average across heads, sum within each span $S\in\{\text{needle},\text{haystack},\text{context}\}$, and average across layers, obtaining sample-level $A_S$, the total mass that $\qstar$ allocates to span $S$ (with $\sum_S A_S = 1$).

\begin{figure}[t]
  \centering
  \includegraphics[width=\columnwidth]{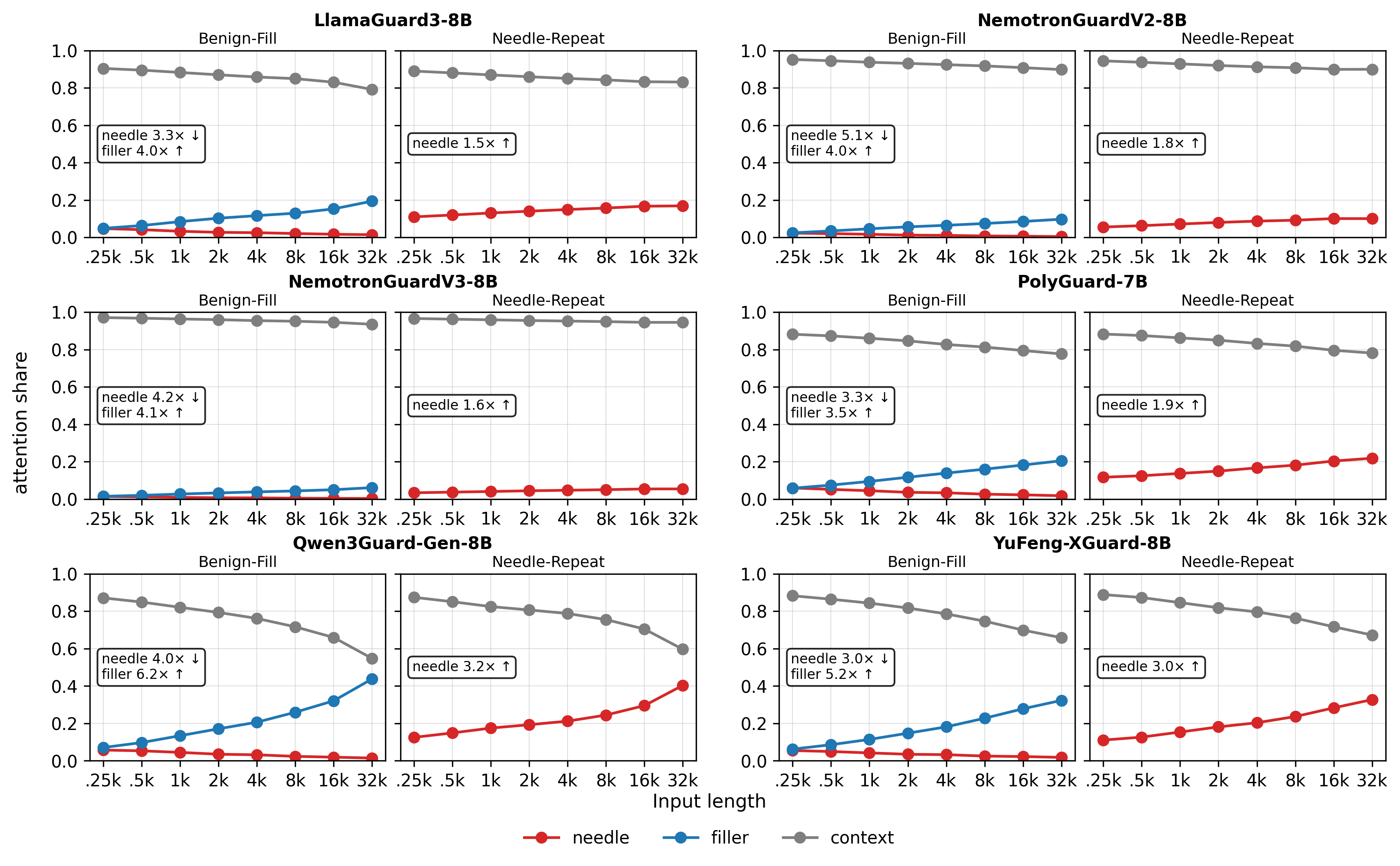}
  \caption{Attention mass of the decision token over the three spans across context lengths (log scale). For each model, left: Benign-Fill, right: Needle-Repeat; the annotation indicates the change from 0.25k to 32k.}
  \label{fig:dilution}
\end{figure}

In Figure~\ref{fig:dilution} (left), the $\Aneedle$ of all six models drops by 3.0--5.1$\times$ from 0.25k to 32k under Benign-Fill, while $\Ahaystack$ rises monotonically and $\Acontext$ slightly decreases in most models. This is a consistent dilution signature: unsafe-needle attention is taken away by the growing haystack rather than by template anchors. The Needle-Repeat counterpart on the right gives the reverse evidence: once dilution is removed, $\Aneedle$ at 32k is higher than at 0.25k, confirming that absolute length itself does not actively steal attention.

\subsubsection{Logit: unsafe margin is compressed}

We define the logit margin at $\qstar$ as
\begin{equation}
\mathrm{margin} = \ell(\mathrm{unsafe}) - \ell(\mathrm{safe}),
\label{eq:margin}
\end{equation}
taking the maximum logit over the unsafe/safe token sets specified by each guardrail's output protocol (e.g.\ \texttt{safe}/\texttt{unsafe} for LlamaGuard3-8B~\cite{inan2023llama}, \texttt{no}/\texttt{yes} for PolyGuard-7B~\cite{kumar2025polyguard}). A positive margin means the argmax is unsafe.

\begin{figure}[t]
  \centering
  \includegraphics[width=\columnwidth]{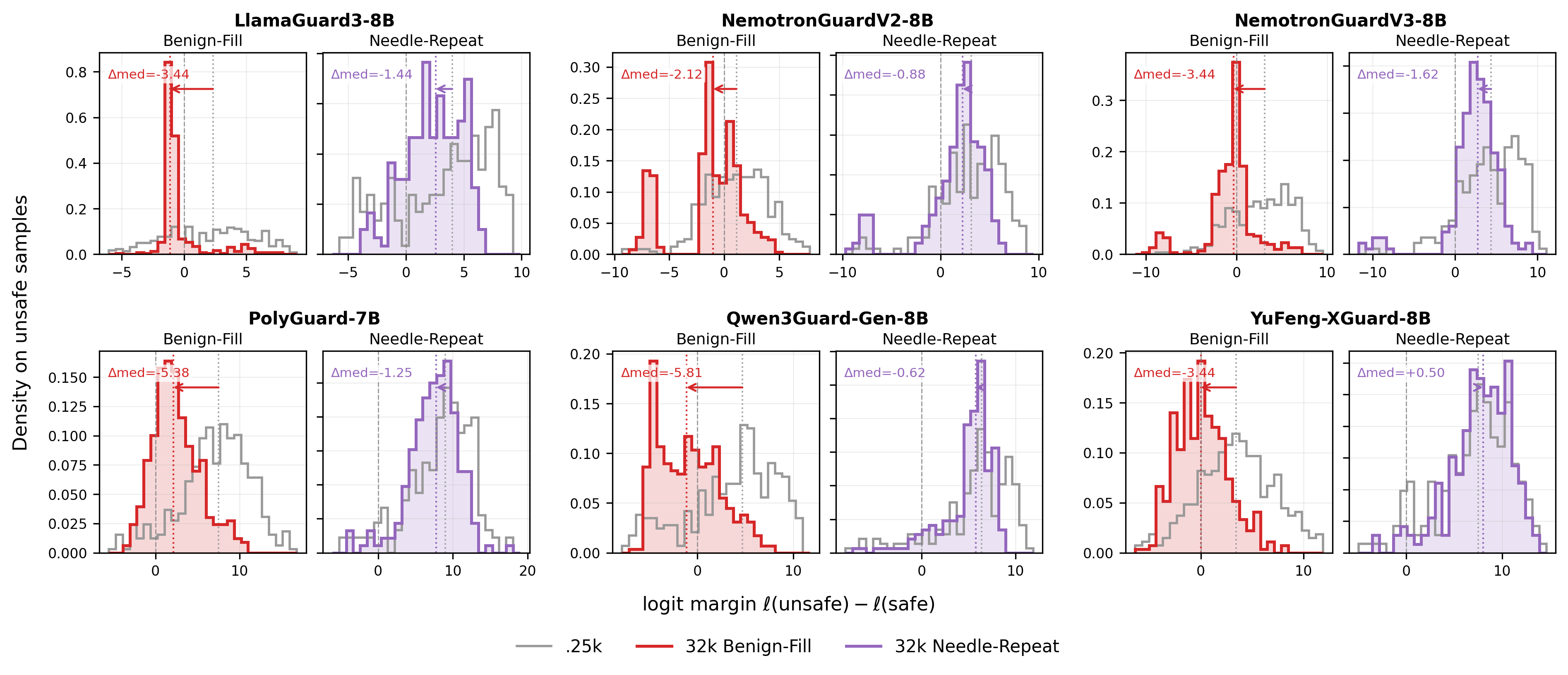}
  \caption{Per-model logit-margin distribution on unsafe samples: 0.25k (gray) vs.\ 32k (red on Benign-Fill, purple on Needle-Repeat); dashed lines mark medians.}
  \label{fig:logit-margin}
\end{figure}

In Figure~\ref{fig:logit-margin} (left), all six Benign-Fill 32k distributions shift left relative to 0.25k, and several cross into the safe ($<\!0$) region; the sample-mean margin is uniformly compressed from 0.25k to 32k. The Needle-Repeat counterpart on the right shows no systematic left shift, with most models stable or slightly increasing across length, fully consistent with the attention-level and behavior-level Needle-Repeat results.

\subsubsection{Mediation consistency}
Beyond directional agreement, we verify that the three layers co-move beyond a common length covariate. On the 48 (model, length) units of Benign-Fill / unsafe, the Pearson partial correlations after controlling for $\log_2 L$ are $+0.65$ (attention$\to$logit), $+0.83$ (logit$\to$behavior), and $+0.56$ (attention$\to$behavior), all with $p<10^{-4}$ (Appendix~\ref{app:mediation}). Together with the Needle-Repeat null (no dilution $\to$ no compression $\to$ no recall drop), this supports proportional dilution as the leading mechanism candidate.

\subsection{Guard-Specialized Retrieval Heads}
\label{sec:retrieval-heads}

The previous section characterizes $\Aneedle$ dilution at the average level. We now decompose it to (layer, head) granularity to see whether the evidence is read uniformly or by a few specialized heads.

\begin{figure}[h!]
  \centering
  \includegraphics[width=\columnwidth]{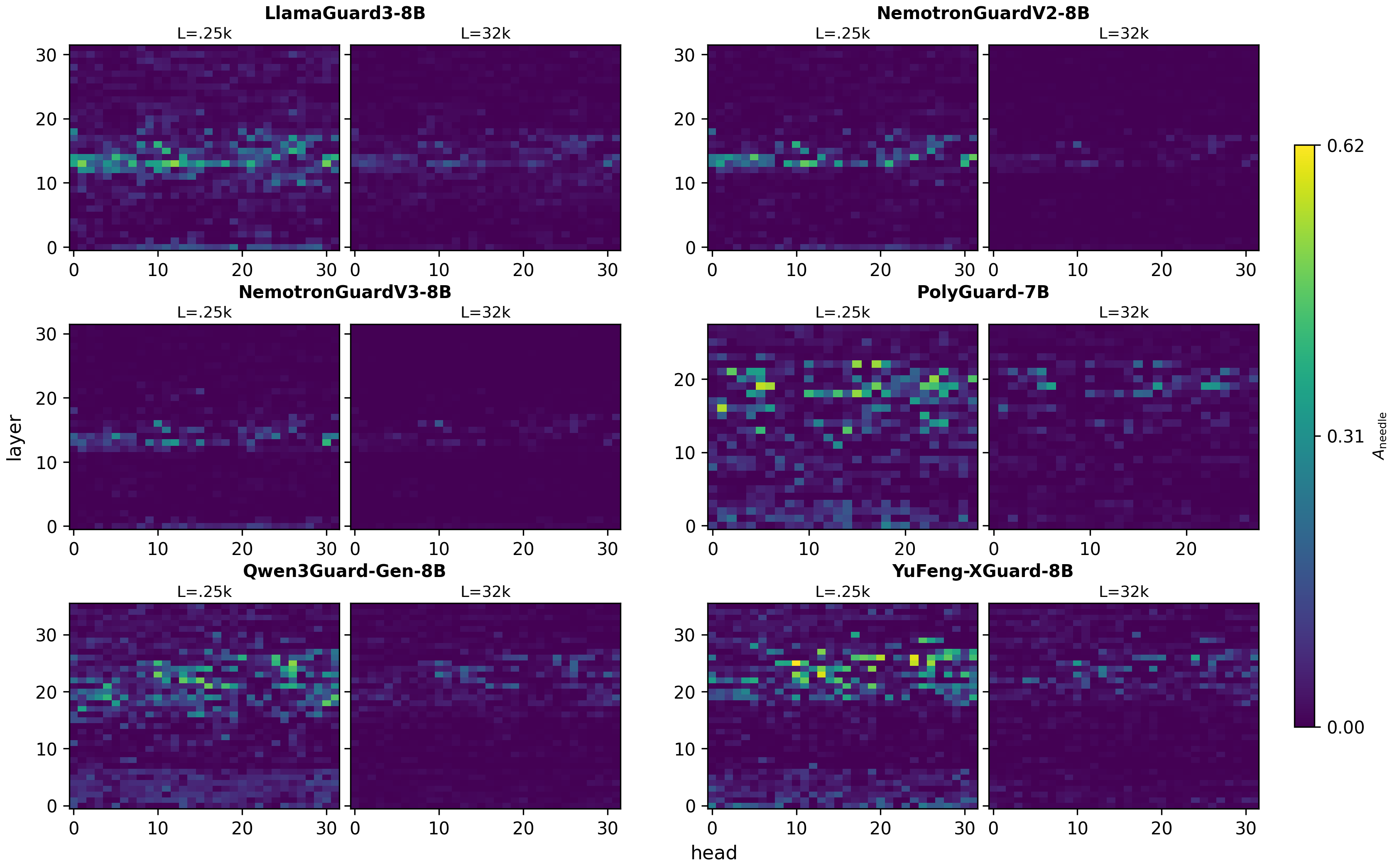}
  \caption{Per-head $\Aneedle^{(\ell,h)}$ heatmaps at $L\!=\!0.25k$ vs.\ $32k$ (shared vmax). Sparse middle-late heads darken at 32k; full sweep in Appendix~\ref{app:headmap}.}
  \label{fig:head-heatmap}
\end{figure}

\begin{figure}[h!]
  \centering
  \includegraphics[width=\columnwidth]{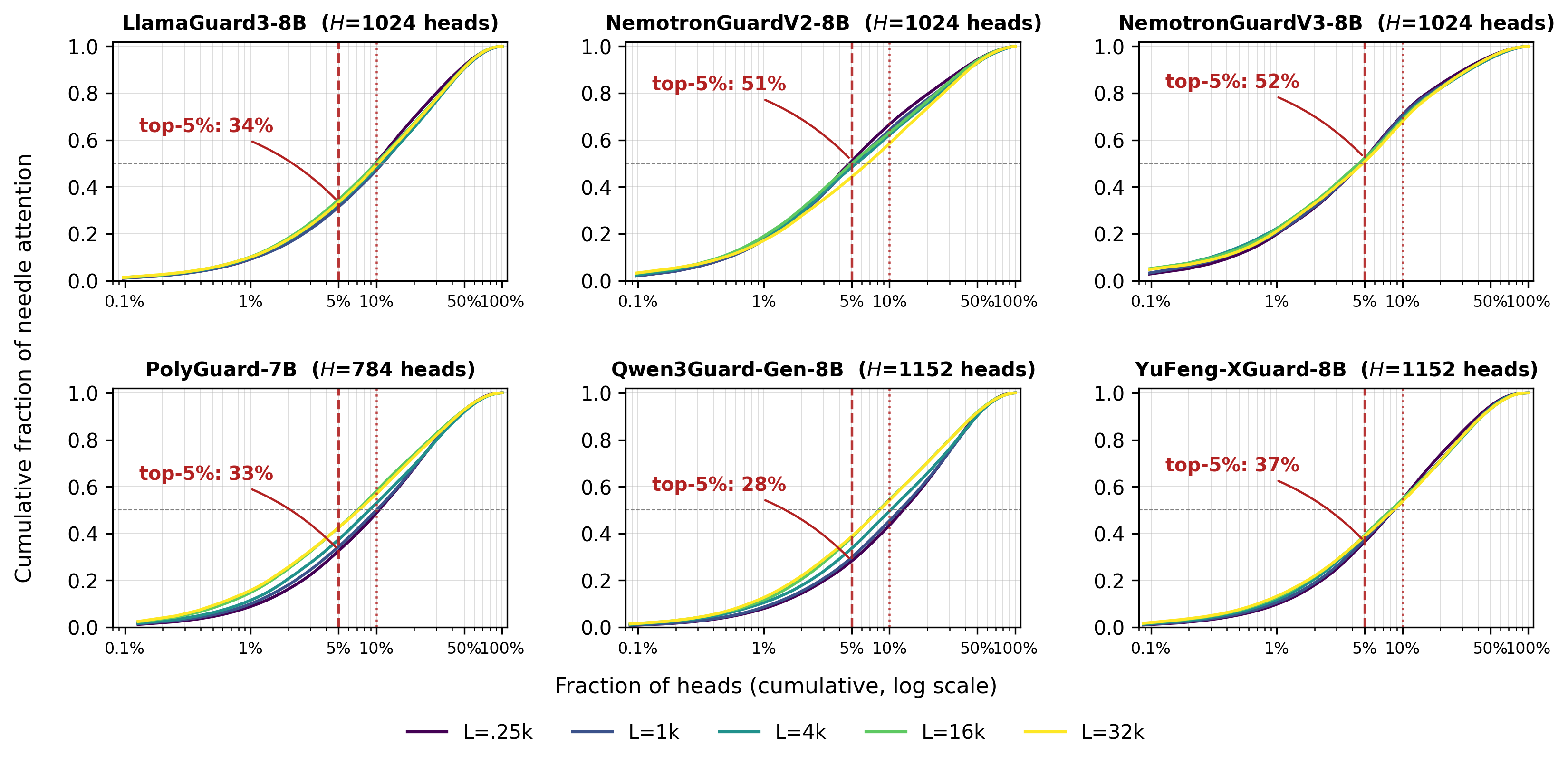}
  \caption{Cumulative needle-attention contribution of the top-$k$ heads. The horizontal axis is the fraction of heads sorted by $\Aneedle$ (descending, log scale); the vertical axis is the cumulative attention share. Top-5\% heads per guardrail contribute $\geq30\%$ at $L\!=\!0.25\text{k}$.}
  \label{fig:head-topk}
\end{figure}

Figures~\ref{fig:head-heatmap} and \ref{fig:head-topk} show that unsafe-needle attention is sparsely concentrated on a few heads. These heads exhibit much higher density on the needle than on the haystack and persist across lengths. Quantitatively, the top-1 head carries 0.77\%--2.84\% of the needle attention; top-5\% heads cumulatively account for 28.3\%--52.4\% across six models (top-10\% reaches 43.6\%--70.9\%).

\begin{figure}[t!]
  \centering
  \includegraphics[width=\columnwidth]{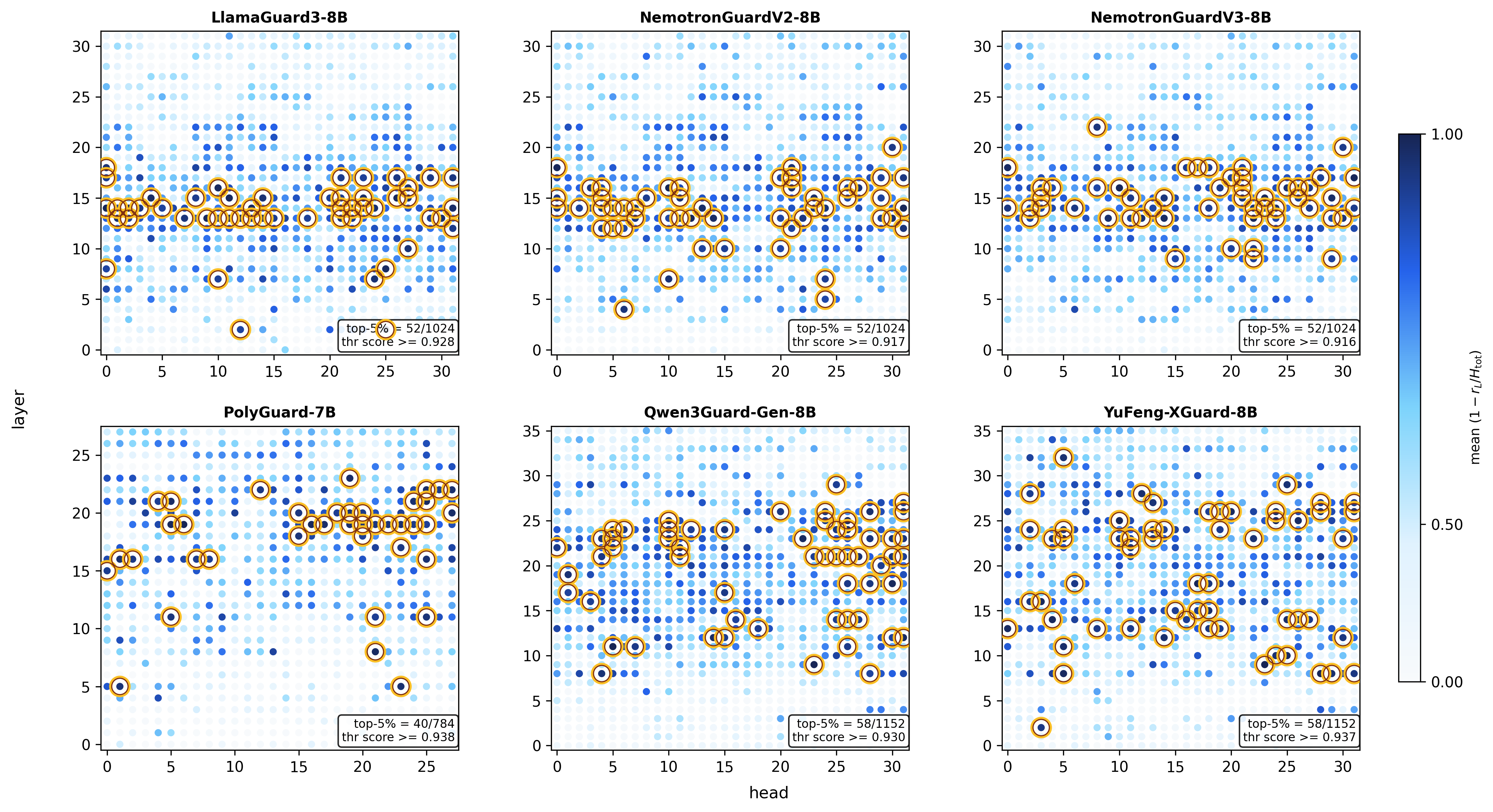}
  \caption{Distribution of $\Hsafety$ on the (layer, head) plane. Background color: the cross-length stability score $\mathcal{S}(l,h)$. Yellow circles: top-5\% by $\mathcal{S}$, falling in the 40\%--70\% middle-late depth.}
  \label{fig:head-selectivity}
\end{figure}

We formalize \emph{needle-selectivity + cross-length stability} as the criterion that defines \emph{guard-specialized retrieval heads} $\Hsafety$. For each (sample, head), we compute the per-token attention density ratio between the needle and the haystack:
\begin{equation}
\sigma^{(l,h)} = \frac{\Aneedle^{(l,h)} / |\text{needle}|}{\Ahaystack^{(l,h)} / |\text{haystack}|},
\label{eq:sigma}
\end{equation}
where $\sigma\gg 1$ indicates a needle-specific head and $\sigma\!\approx\!1$ indicates a broadcast head. For each guardrail and each $L\in\mathcal{L}$, we rank heads by $\sigma$ in descending order ($r_L(l,h)$), and define the cross-length stability score
\begin{equation}
\mathcal{S}(l,h) = \frac{1}{|\mathcal{L}|} \sum_{L\in\mathcal{L}}\left(1 - \frac{r_L(l,h)}{H_{\mathrm{tot}}}\right).
\label{eq:stability}
\end{equation}
We take the top-$K\%$ heads of each guardrail by $\mathcal{S}$ as $\Hsafety^{(K\%)}$. Throughout this section, we set $K=5$. Figure~\ref{fig:head-selectivity} shows that $\Hsafety$ concentrates in the 40\%--70\% middle-late depth and is sparse in width.

\paragraph{Partial specificity vs.\ base.} A base--guard control on six pairs (Appendix~\ref{app:base-vs-guard}) shows that the top-5\% heads of $\Hsafety$ exhibit \emph{partial specificity}: the cross-length rank correlation between guard and base remains $\rho\!\geq\!0.74$ (the head identity largely overlaps with what already exists in the base), yet at 32k the guard's top-5\% share is $+7.9\%$ to $+19.9\%$ higher than the base's, and per-head selectivity $\log_{10}\sigma$ is 1.19$\times$--3.95$\times$ larger in median. Guard fine-tuning, therefore, does not create new retrieval pathways but amplifies the per-head magnitude and cross-length concentration of needle-specific heads on the existing retrieval topology of the base.

\subsection{Sufficiency Interventions}
\label{sec:intervention}

Two positive interventions verify the chain's operability. \emph{Token-level:} restricting visible context to the unsafe needle (\texttt{needle-only}) lifts $\Ru$ from the 47.8\% \texttt{full} baseline to 70.2\% ($+$22.5\%), while keeping an equal amount of \emph{random} haystack instead drops it to 24.0\% ($-$23.8\%) -- recovery comes from preserving unsafe evidence, not from shortening the input. \emph{Head-level:} post-hoc amplifying $\qstar$'s attention to the needle on $\Hsafety$ with multipliers $\alpha\!\in\!\{2,4,8\}$ monotonically raises $\Ru$ by $+$1.5\%, $+$3.8\%, $+$6.4\% on all six guardrails, while the same operation on an equal number of random heads stays within $[-0.3\%, -0.1\%]$. Token-level evidence preservation and head-level evidence amplification both contribute functionally to long-context safety judgment under dilution. The full intervention design, formulas, and dose-response curves are deferred to Appendix~\ref{app:intervention}.

\section{Training-Free Mitigations}
\label{sec:method}

Targeting RQ3, this section turns the analysis into two training-free mitigations -- \textbf{Chunked Detection} (CD), which chunks the input to raise the local needle share, and \textbf{Attention-Head Sharpening} (AHS), which applies a softmax temperature on $\Hsafety$ -- and a deployment protocol, \textbf{Context-Aware Hyperparameter Routing} (CAHR), that unifies their hyperparameter selection.

\subsection{Chunked Detection (CD)}
\label{sec:cd}

Section~\ref{sec:degradation-cause} localizes the dominant cause to dilution. The simplest counter-intervention is to chunk the long input so that the local needle share returns from the diluted regime to the dense regime. CD splits the input $x$ into $N=\lceil |x|/W\rceil$ non-overlapping segments of $W$ words, queries the guardrail on each segment independently, and aggregates by OR: any segment marked unsafe yields an unsafe verdict. The tunable hyperparameter is $W\in\{64, 128, 256, 512\}$ words.

\subsection{Attention-Head Sharpening (AHS)}
\label{sec:ahs}

AHS reparameterizes the hard amplification of Section~\ref{sec:intervention} -- which requires needle annotations and is not deployable -- as a softmax-temperature sharpening ($\tau<1$) on $\Hsafety$ that no longer needs labels. The feasibility is directly supported by Section~\ref{sec:retrieval-heads}: the needle-vs-haystack selectivity $\sigma\gg 1$ on $\Hsafety$ holds at every length, so dilution compresses the absolute magnitude of needle attention but not its relative ranking. Temperature sharpening amplifies the pre-softmax logit gap and converts the residual relative advantage into absolute attention mass, thereby achieving label-free directed amplification. Partial specificity offers two layers of safety. First, AHS sharpens heads that fine-tuning has already strengthened, thereby avoiding triggering out-of-distribution behavior unseen in the base. Second, selecting the sparse top-5\% subset by selectivity naturally excludes attention-sink heads (whose weights are routed to BOS/system tokens, $\Aneedle\!\approx\!0$), avoiding the sink-collapse side effect of network-wide sharpening.

Concretely, AHS scales the pre-softmax logits of each $h\!\in\!\Hsafety$ by temperature $\tau\!\in\!(0,1)$ during forward computation while leaving the other heads intact:
\begin{equation}
\tilde A^{(h)}_{ij} = \mathrm{softmax}_{j}\!\left(\frac{q_{i}^{(h)\top} k_{j}^{(h)}}{\tau\sqrt{d_{k}}}\right).
\label{eq:ahs}
\end{equation}
$\Hsafety$ is reused directly from Section~\ref{sec:retrieval-heads}; the tunable hyperparameters are $K\!\in\!\{1,5,10\}$ and $\tau\!\in\!\{0.1, 0.3, 0.5, 0.7\}$.

\subsection{Context-Aware Hyperparameter Routing (CAHR)}
\label{sec:cahr}

The length-wise sweep in Appendix~\ref{app:cahr-sweep} shows that \emph{the optimal configuration of a guardrail differs across context lengths}, so any fixed configuration is suboptimal in some range. CAHR formalizes configuration selection as a routing table $\pi_{M,\mathcal{F}}(L, t)$ over length$\times$audit side; the table is fitted on SafetyNIAH and frozen at deployment.

\paragraph{Offline fitting.} For each guardrail $M$ and method family $\mathcal{F}\!\in\!\{\mathrm{CD}, \mathrm{AHS}\}$, SafetyNIAH is split into $8\!\times\!2$ cells indexed by $(L, t)\!\in\!\mathcal{L}\!\times\!\{\mathrm{prompt}, \mathrm{response}\}$; on each cell we record $\pi_{M,\mathcal{F}}(L, t)=\arg\max\Funsafe$ over all candidates, including the no-operation option $\varnothing$.

\paragraph{Online inference.} For each input $x$ with known audit side $t$, CAHR first runs the base prediction, length bucket, and routing lookup,
\begin{equation}
\begin{aligned}
y_{\mathrm{base}} &\leftarrow M(x), \\
\hat L &\leftarrow \mathrm{bucket}(|x|), \\
\hat\theta &\leftarrow \pi_{M,\mathcal{F}}(\hat L, t),
\end{aligned}
\label{eq:cahr-online}
\end{equation}
and returns $y_{\mathrm{base}}$ if $y_{\mathrm{base}}=\mathrm{unsafe}$ (base-cascade gating, preventing chunk voting or attention sharpening from flipping a correctly detected unsafe case back to safe) or $\hat\theta=\varnothing$; otherwise it returns $M_{\hat\theta}(x)$. All three features (length, audit side, base prediction) are deployment-observable, and Appendix~\ref{app:cahr-ablation} confirms that none of them is redundant.

\begin{table*}[t!]
  \centering
  \resizebox{\textwidth}{!}{%
  \begin{tabular}{lcccccc}
    \toprule
    \multirow{2}{*}{Model / Method} & \multicolumn{1}{c}{\emph{In-domain synthetic}} & \multicolumn{2}{c}{\emph{OOD input-side long-context attacks}} & \multicolumn{2}{c}{\emph{OOD output-side long outputs}} & \multirow{2}{*}{Avg.\ $\Delta$} \\
    \cmidrule(lr){2-2} \cmidrule(lr){3-4} \cmidrule(lr){5-6}
     & SafetyNIAH $\Funsafe\uparrow$ & MSJ $\Ru\uparrow$ & NINJA $\Ru\uparrow$ & Qwen3G-long $\Funsafe\uparrow$ & RShield-long $\Funsafe\uparrow$ & \\
    \midrule
    \textbf{LlamaGuard3-8B}      & 50.00 & 75.00 & 60.31 & 49.34 & 44.93 & -- \\
    \quad + CAHR-CD              & 77.42 & 100.00 & 99.27 & 67.54 & 67.16 & $+$26.36 \\
    \quad + CAHR-AHS             & 63.50 & 80.94 & 80.73 & 57.80 & 53.92 & $+$11.46 \\
    \hdashline
    \textbf{NemotronGuardV2-8B}  & 48.38 & 75.62 & 35.21 & 16.47 & 19.60 & -- \\
    \quad + CAHR-CD              & 79.60 & 100.00 & 67.71 & 78.37 & 69.52 & $+$39.98 \\
    \quad + CAHR-AHS             & 75.01 & 76.56 & 53.02 & 69.11 & 61.99 & $+$28.08 \\
    \hdashline
    \textbf{NemotronGuardV3-8B}  & 53.84 & 80.62 & 53.54 & 13.93 & 11.60 & -- \\
    \quad + CAHR-CD              & 82.75 & 100.00 & 70.00 & 84.43 & 76.42 & $+$40.01 \\
    \quad + CAHR-AHS             & 75.59 & 90.00 & 66.56 & 63.31 & 51.88 & $+$26.76 \\
    \hdashline
    \textbf{PolyGuard-7B}        & 80.77 & 99.69 & 75.73 & 77.46 & 78.07 & -- \\
    \quad + CAHR-CD              & 82.83 & 100.00 & 75.73 & 76.92 & 75.85 & $-$0.08 \\
    \quad + CAHR-AHS             & 82.11 & 100.00 & 80.62 & 78.99 & 76.45 & $+$1.29 \\
    \hdashline
    \textbf{Qwen3Guard-Gen-8B (strict)} & 80.29 & 100.00 & 92.50 & 79.71 & 83.62 & -- \\
    \quad + CAHR-CD              & 85.19 & 100.00 & 96.35 & 79.71 & 83.62 & $+$1.75 \\
    \quad + CAHR-AHS             & 83.25 & 100.00 & 95.73 & 80.58 & 84.34 & $+$1.56 \\
    \hdashline
    \textbf{YuFeng-XGuard-8B}    & 75.85 & 10.62$^{\dagger}$ & 72.71 & 73.99 & 83.68 & -- \\
    \quad + CAHR-CD              & 84.39 & 100.00 & 78.44 & 83.84 & 83.59 & $+$22.68 \\
    \quad + CAHR-AHS             & 78.97 & 31.56 & 74.38 & 78.66 & 84.32 & $+$6.21 \\
    \midrule
    \rowcolor{gray!15}
    \textbf{six-guard mean}      & 64.86 & 73.59 & 65.00 & 51.82 & 53.58 & -- \\
    \rowcolor{gray!15}
    \quad + CAHR-CD              & 82.03 & 100.00 & 81.25 & 78.47 & 76.03 & $+$21.79 \\
    \rowcolor{gray!15}
    \quad + CAHR-AHS             & 76.41 & 79.84 & 75.17 & 71.41 & 68.81 & $+$12.56 \\
    \bottomrule
  \end{tabular}}
  \caption{Cross-distribution performance on five long-context benchmarks; ``Avg.\ $\Delta$'' is the mean improvement over the original. $^{\dagger}$ YuFeng-XGuard-8B's low MSJ baseline reflects 88.1\% invalid parses hijacked by the in-context demonstrations, not detection collapse.}
  \label{tab:main-method}
\end{table*}

\section{Evaluation of Mitigations}
\label{sec:eval}

This section continues to answer RQ3 by checking whether the two methods consistently restore detection across long-context distributions without significantly inflating false alarms on safe samples.

\subsection{Experimental Setup}

\paragraph{Benchmarks.} We evaluate on five test sets across three categories.

\emph{In-domain synthetic}: the \textbf{SafetyNIAH} main benchmark (30{,}400 samples). We split SafetyNIAH at the needle level into a 7:3 dev/test split. The routing table is fitted on the dev split; SafetyNIAH scores are computed on the test split. Appendix~\ref{app:cahr-devtest} reports 5-fold cross-validation.

\emph{OOD input-side long-context attacks}: \textbf{MSJ}~\cite{anil2024many} (320 samples) prepends multi-turn harmful demonstrations to a target query, and \textbf{NINJA}~\cite{shah2025jailbreaking} (960 samples) embeds harmful instructions into LLM-generated benign haystacks with front/middle/end positions; both reuse 80 HarmBench-val harmful goals as targets.

\emph{OOD output-side long outputs}: \textbf{Qwen3G-long} (1{,}059 samples) and \textbf{RShield-long} (1{,}866 samples) extend Qwen3GuardTest~\cite{zhao2025qwen3guard} and ReasoningShieldTest~\cite{li2025reasoningshield} (avg.\ $\leq\!2$k tokens) by treating each original response as a needle and filling Wikipedia chunks up to 1k--16k, mirroring the SafetyNIAH response-side construction.

\paragraph{Metrics.} For benchmarks containing both unsafe and safe samples (SafetyNIAH, Qwen3G-long, RShield-long), we report $\Funsafe$ as the primary metric, which jointly constrains precision and recall and prevents inflating recall by safe$\to$unsafe misclassification. For the attack-only benchmarks (MSJ, NINJA), we report the unsafe recall $\Ru$.

\begin{figure}[h!]
  \centering
  \includegraphics[width=\columnwidth]{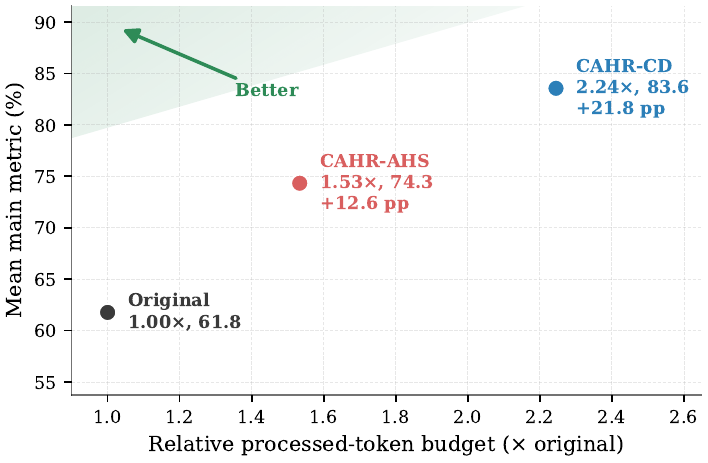}
  \caption{Effect--cost trade-off of CAHR. Horizontal: relative processed-token budget ($\times$ original); vertical: six-guard mean of the five primary metrics.}
  \label{fig:cost-effect}
\end{figure}

\subsection{Results}

\paragraph{Consistent gains across benchmarks.} On the six-guard mean (Table~\ref{tab:main-method}), CAHR-CD and CAHR-AHS surpass the base on every primary column by $+$21.79\% and $+$12.56\% on average. Routing tables are fitted on the dev split of SafetyNIAH only, yet the consistent gains on the four frozen OOD benchmarks (MSJ, NINJA, Qwen3G-long, RShield-long) suggest these are not surface patterns of synthetic needle--haystack pairs.

\paragraph{Length-wise recovery.} By length, the base drops monotonically on every benchmark (SafetyNIAH 82.7$\to$40.2 at 0.25k$\to$32k; MSJ 93.3$\to$39.4 at 4k$\to$64k), while the CAHR-CD and CAHR-AHS curves flatten (CAHR-CD on SafetyNIAH: 85.0$\to$79.4, $\sim\!$ 1/8 of the base drop). The gains target the long-context dimension specifically; per-length curves are in Appendix~\ref{app:length-gain}.

\paragraph{Effect--cost trade-off.} CAHR-CD attains the larger gain at higher token cost, though the gain shrinks toward zero on already-robust guards (PolyGuard $-0.08$, Qwen3Guard $+1.75$), while CAHR-AHS gives milder gains ($[+1.29, +28.08]$) at a cost close to the base; the two fit different deployment budgets (Figure~\ref{fig:cost-effect}).

\section{Generalization Across Filler Domains}
\label{sec:cross-domain}

Sections~\ref{sec:degradation} and~\ref{sec:eval} establish two findings on SafetyNIAH, whose haystack is drawn from English Wikipedia: unsafe recall degrades monotonically with length, and CAHR-CD/CAHR-AHS consistently recover it. Here we check whether both findings persist when the filler is replaced by natural text from other domains and languages.

\paragraph{Construction.} We uniformly sample 800 needles from the SafetyNIAH needle pool, balanced across length, label, and source benchmark, and hold the needle/position/length grid fixed as in Section~\ref{sec:safetyniah}. We replace the English-Wikipedia haystack with three natural sources:
\begin{itemize}
  \item \emph{conversation} -- real multi-turn chats from WildChat~\cite{zhao2024wildchat};
  \item \emph{code} -- multi-language source code from The Stack~\cite{kocetkov2022stack};
  \item \emph{Chinese} -- non-English text from Chinese Wikipedia,\footnote{\url{https://huggingface.co/datasets/wikimedia/wikipedia/viewer/20231101.zh}} a language that has dedicated safety moderators~\cite{chen2025libra}.
\end{itemize}
This yields 2{,}400 samples, which we evaluate on the six guardrails from Section~\ref{sec:mechanism} under the base guard, CAHR-CD, and CAHR-AHS.

\paragraph{Both conclusions hold across domains.} Table~\ref{tab:cross-domain-recall} shows that base unsafe recall degrades monotonically with length in every domain (six-guard mean $\Delta$ from 0.25k to 32k: $-41.3\%$ conversation, $-34.7\%$ code, $-33.0\%$ Chinese), matching the Wikipedia-filler result of Section~\ref{sec:degradation-main}. Table~\ref{tab:cross-domain-f1} shows that CAHR-CD and CAHR-AHS raise $\Funsafe$ over the base in every domain, with CD giving the larger gain ($+12.1$ to $+24.0$) and AHS the cheaper one ($+8.6$ to $+12.0$), mirroring the effect--cost ordering of Section~\ref{sec:eval}. These results indicate that both the degradation and the mitigation gains generalize beyond English-Wikipedia filler to conversational, code, and non-English text.

\begin{table}[t]
\centering
\resizebox{\columnwidth}{!}{%
\begin{tabular}{lcccccc}
\toprule
Domain & 0.25k & 0.5k & 2k & 8k & 32k & $\Delta\Ru$ \\
\midrule
\emph{conversation} & 68.7 & 59.3 & 44.0 & 30.3 & 27.3 & $-$41.3 \\
\emph{code} & 83.7 & 77.0 & 66.0 & 57.0 & 49.0 & $-$34.7 \\
\emph{Chinese} & 71.7 & 63.7 & 51.0 & 44.3 & 38.7 & $-$33.0 \\
\bottomrule
\end{tabular}}
\caption{Unsafe recall $\Ru$ by filler domain and context length (six-guard mean, \%).}
\label{tab:cross-domain-recall}
\end{table}

\begin{table}[h!]
\centering
\resizebox{0.9\columnwidth}{!}{%
\begin{tabular}{lccc}
\toprule
Domain & base & CAHR-CD & CAHR-AHS \\
\midrule
\emph{conversation} & 47.0 & 71.0 ($+$24.0) & 59.0 ($+$12.0) \\
\emph{code} & 67.9 & 80.0 ($+$12.1) & 76.5 ($+$8.6) \\
\emph{Chinese} & 57.4 & 74.1 ($+$16.7) & 67.9 ($+$10.5) \\
\bottomrule
\end{tabular}}
\caption{$\Funsafe$ by filler domain: base vs.\ CAHR-CD vs.\ CAHR-AHS (six-guard mean, \%, over all lengths).}
\label{tab:cross-domain-f1}
\end{table}

\section{Conclusion}
\label{sec:conclusion}

We present LongGuard, a unified framework for evaluating, analyzing, and mitigating long-context guardrail failure. SafetyNIAH shows that across 15 guards, unsafe recall drops by over 50\% on average, driven by proportional dilution of the unsafe needle rather than absolute length. A three-layer attention$\to$logit$\to$behavior analysis localizes the mechanism to a sparse set of guard-specialized retrieval heads with \emph{partial specificity} to their base models. Two training-free methods (CD, AHS) under CAHR routing then improve five cross-distribution benchmarks by $+$21.79\% and $+$12.56\% on average. Both the degradation and the mitigation effects also generalize across domains and languages. We hope these results push guardrail training and evaluation to treat context length as a first-class axis.

\section*{Limitations}

This paper primarily bases its experimental analysis on controllably synthesized samples. Future research will further develop long-context safety datasets that more faithfully reflect real-world scenarios and are collected from naturally occurring interactions, enabling more comprehensive testing and analysis of the phenomena under investigation. In addition, the mitigation strategies studied in this paper are deliberately positioned as training-free, inference-time methods; approaches that require fine-tuning the guard model lie outside the scope of this paper.

\section*{Ethical Considerations}

Our findings are double-edged: they can inform guardrail improvement, but they could also be exploited by attackers. SafetyNIAH is a diagnostic stress test, not a training corpus: needles inherit 17 public benchmarks (Appendix~\ref{app:data}) under their original licenses, haystacks are guardrail-filtered, and access is restricted to safety research with upstream takedowns honored.

\section*{Acknowledgements}

This work is supported by the National Natural Science Foundation of China (No. U24A20335).

\bibliography{longguard}

\clearpage

\appendix

\begin{figure*}[t!]
  \hspace*{3em}\includegraphics[width=\dimexpr\textwidth-1em\relax]{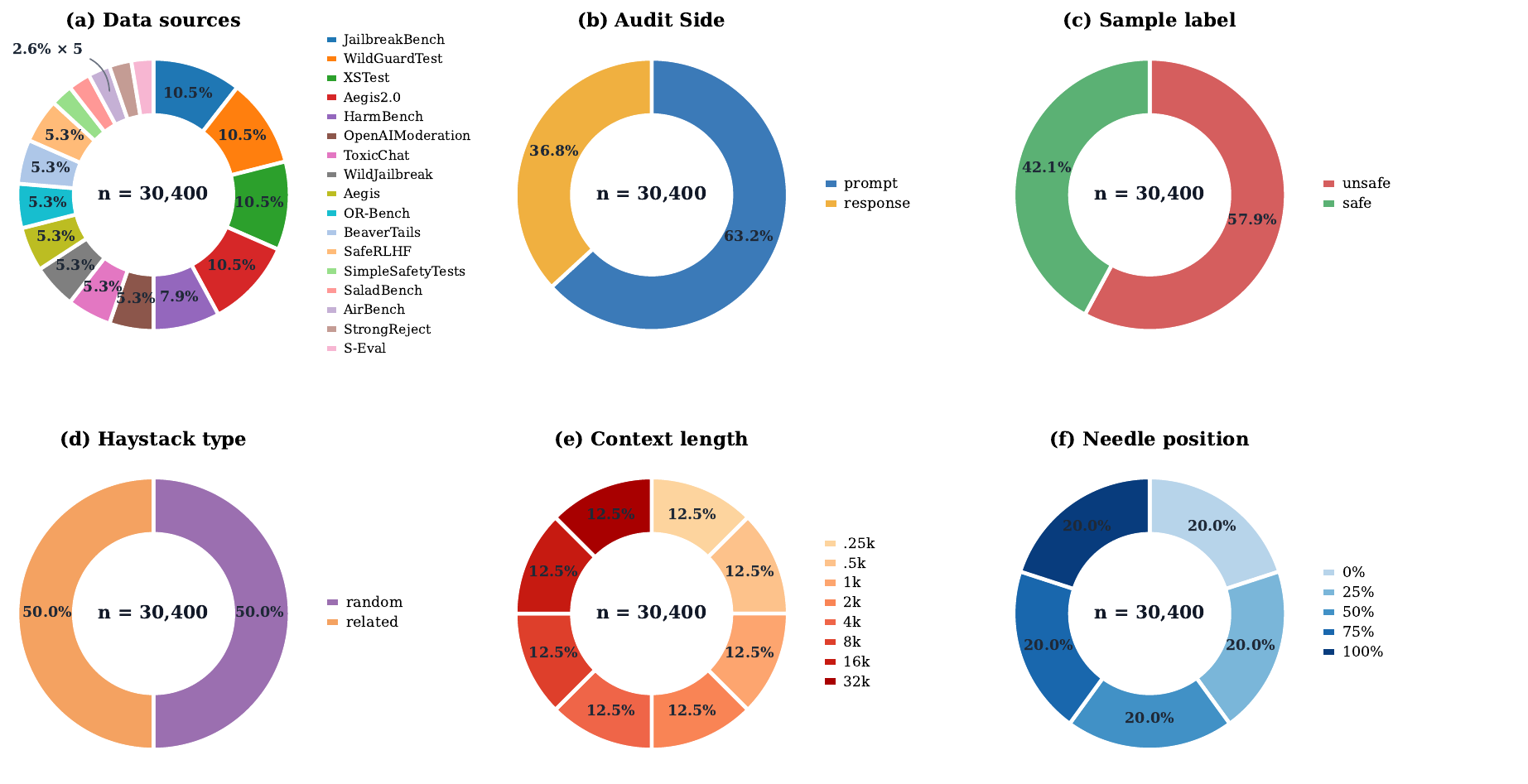}
  \caption{Composition of the SafetyNIAH main benchmark.}
  \label{fig:benchmark-dist}
\end{figure*}

\begin{table*}[h!]
\centering
\resizebox{\textwidth}{!}{%
\begin{tabular}{lcccccc}
\toprule
Source benchmark & Use & Audit side & \#Samples & Label ratio & Avg.\ length (words) & License \\
\midrule
OpenAIModeration~\cite{markov2023holistic} & SafetyNIAH needle pool & input & 1{,}680 & safe 68.93\% / unsafe 31.07\% & prompt 111.0 & MIT \\
ToxicChat~\cite{lin2023toxicchat} & SafetyNIAH needle pool & input & 5{,}083 & safe 92.88\% / unsafe 7.12\% & prompt 28.2 & CC-BY-NC-4.0 \\
SimpleSafetyTests~\cite{vidgen2023simplesafetytests} & SafetyNIAH needle pool & input & 100 & unsafe 100.00\% & prompt 11.4 & CC-BY-2.0 \\
WildJailbreak~\cite{jiang2024wildteaming} & SafetyNIAH needle pool & input & 2{,}210 & safe 9.50\% / unsafe 90.50\% & prompt 120.6 & ODC-BY \\
SaladBench~\cite{li2024salad} & SafetyNIAH needle pool & input & 26{,}318 & unsafe 100.00\% & prompt 94.0 & Apache-2.0 \\
AirBench~\cite{zeng2025air} & SafetyNIAH needle pool & input & 5{,}694 & unsafe 100.00\% & prompt 120.6 & CC-BY-4.0 \\
StrongReject~\cite{souly2024strongreject} & SafetyNIAH needle pool & input & 313 & unsafe 100.00\% & prompt 25.9 & MIT \\
Aegis~\cite{ghosh2024aegis} & SafetyNIAH needle pool & input & 1{,}199 & safe 46.96\% / unsafe 53.04\% & prompt 130.9 & CC-BY-4.0 \\
OR-Bench~\cite{cui2024or} & SafetyNIAH needle pool & input & 1{,}974 & safe 66.82\% / unsafe 33.18\% & prompt 17.0 & CC-BY-4.0 \\
S-Eval~\cite{yuan2025s} & SafetyNIAH needle pool & input & 110{,}000 & unsafe 100.00\% & prompt 182.1 & CC-BY-NC-SA-4.0 \\
\multirow{2}{*}{JailbreakBench~\cite{chao2024jailbreakbench}} & \multirow{2}{*}{SafetyNIAH needle pool} & input  & 200      & safe 50.00\% / unsafe 50.00\% & prompt 12.9                  & \multirow{2}{*}{MIT}       \\
                                &                                          & output & 300      & safe 63.33\% / unsafe 36.67\% & prompt 79.7 / response 99.4   &                            \\
\multirow{2}{*}{WildGuardTest~\cite{han2024wildguard}}  & \multirow{2}{*}{SafetyNIAH needle pool} & input  & 1{,}699  & safe 55.62\% / unsafe 44.38\% & prompt 75.4                  & \multirow{2}{*}{ODC-BY}    \\
                                &                                          & output & 1{,}709  & safe 83.38\% / unsafe 16.62\% & prompt 75.1 / response 272.0  &                            \\
\multirow{2}{*}{XSTest~\cite{rottger2024xstest}}         & \multirow{2}{*}{SafetyNIAH needle pool} & input  & 446      & safe 55.83\% / unsafe 44.17\% & prompt 8.4                   & \multirow{2}{*}{ODC-BY}    \\
                                &                                          & output & 446      & safe 82.51\% / unsafe 17.49\% & prompt 8.4 / response 139.4   &                            \\
\multirow{2}{*}{HarmBench~\cite{mazeika2024harmbench}}      & \multirow{2}{*}{SafetyNIAH needle pool} & input  & 320      & unsafe 100.00\%                & prompt 64.8                  & \multirow{2}{*}{MIT}       \\
                                &                                          & output & 602      & safe 54.65\% / unsafe 45.35\% & prompt 160.2 / response 218.1 &                            \\
\multirow{2}{*}{Aegis2.0~\cite{ghosh2025aegis2}}       & \multirow{2}{*}{SafetyNIAH needle pool} & input  & 1{,}964  & safe 46.08\% / unsafe 53.92\% & prompt 44.9                  & \multirow{2}{*}{CC-BY-4.0} \\
                                &                                          & output & 813      & safe 51.54\% / unsafe 48.46\% & prompt 12.9 / response 115.7  &                            \\
BeaverTails~\cite{ji2023beavertails} & SafetyNIAH needle pool & output & 3{,}021 & safe 42.63\% / unsafe 57.37\% & prompt 13.1 / response 59.8 & CC-BY-NC-4.0 \\
SafeRLHF~\cite{ji2025pku} & SafetyNIAH needle pool & output & 16{,}422 & safe 47.69\% / unsafe 52.31\% & prompt 22.2 / response 86.1 & CC-BY-NC-4.0 \\
Qwen3GuardTest~\cite{zhao2025qwen3guard} & OOD long output & output & 1{,}059 & safe 46.27\% / unsafe 53.73\% & prompt 13.2 / response 681.0 & CC-BY-NC-4.0 \\
ReasoningShieldTest~\cite{li2025reasoningshield} & OOD long output & output & 2{,}200 & safe 56.27\% / unsafe 28.55\% / ambig.\ 15.18\% & prompt 83.5 / response 809.6 & CC-BY-NC-4.0 \\
\bottomrule
\end{tabular}}
\caption{Source benchmarks used in this paper.}
\label{tab:source-data}
\end{table*}

\begin{table*}[h!]
\centering
\resizebox{\textwidth}{!}{%
\begin{tabular}{lccccccc}
\toprule
\multirow{2}{*}{Benchmark} & \multirow{2}{*}{\#Samples} & \multirow{2}{*}{\#Safe / \#Unsafe} & \multirow{2}{*}{\#Prompt / \#Response} & \multicolumn{4}{c}{Length (words)} \\
\cmidrule(lr){5-8}
 & & & & Avg. & Median & p90 & Max \\
\midrule
SafetyNIAH (Benign-Fill) & 30{,}400 & 12{,}800 / 17{,}600 & 19{,}200 / 11{,}200 & 8{,}160 & 3{,}072 & 32{,}768 & 32{,}768 \\
SafetyNIAH (Needle-Repeat) & 8{,}800 & 0 / 8{,}800 & 6{,}000 / 2{,}800 & 8{,}160 & 3{,}072 & 32{,}768 & 32{,}768 \\
MSJ & 320 & 0 / 320 & 320 / 0 & 26{,}278 & 21{,}120 & 56{,}217 & 58{,}004 \\
NINJA & 960 & 0 / 960 & 960 / 0 & 4{,}558 & 3{,}502 & 10{,}232 & 10{,}403 \\
Qwen3G-long & 1{,}059 & 490 / 569 & 0 / 1{,}059 & 6{,}264 & 4{,}084 & 16{,}373 & 16{,}381 \\
RShield-long & 1{,}866 & 1{,}238 / 628 & 0 / 1{,}866 & 6{,}319 & 4{,}073 & 16{,}363 & 16{,}381 \\
\bottomrule
\end{tabular}}
\caption{Statistics of the six long-context evaluation sets.}
\label{tab:final-data}
\end{table*}

\section{Benchmark Statistics}
\label{app:data}

Figure~\ref{fig:benchmark-dist} shows the composition of the SafetyNIAH main benchmark over context length, needle position, audit side, and haystack type. Tables~\ref{tab:source-data} and~\ref{tab:final-data} summarize the raw candidate sources and the six long-context evaluation sets, respectively.

\section{Full Model List}
\label{app:models}

Table~\ref{tab:models} lists the 15 open-source safety guardrails benchmarked in this paper. Inference is deterministic (temperature $0$). The strict-filtering subset $\mathcal{G}_{\mathrm{filter}}$ used to construct the neutral haystack pool (Section~\ref{sec:safetyniah}) and the six full-attention guardrails used for mechanistic analysis (Section~\ref{sec:mechanism}) are both drawn from this list.

\begin{table*}[t]
\centering
\resizebox{\textwidth}{!}{%
\begin{tabular}{lcccccc}
\toprule
Model & Size & Backbone & Detection & Output format & License \\
\midrule
LlamaGuard3~\cite{inan2023llama} & 8B & Llama-3.1-8B & input+output & safe/unsafe & Llama 3.1 Community License \\
LlamaGuard4~\cite{inan2023llama} & 12B & Llama-4-Scout & input+output & safe/unsafe & Llama 4 Community License \\
WildGuard~\cite{han2024wildguard} & 7B & Mistral-7B-v0.3 & input+output+refusal & safe/unsafe & Apache-2.0 \\
NemotronGuardV2~\cite{ghosh2025aegis2} & 8B & Llama-3.1-8B-Instruct & input+output & safe/unsafe & NVIDIA OML + Llama 3.1 \\
NemotronGuardV3~\cite{ghosh2025aegis2} & 8B & Llama-3.1-8B-Instruct & input+output & safe/unsafe & NVIDIA OML + Llama 3.1 \\
NemotronReasoning~\cite{sreedhar2025safety} & 4B & Gemma-3-4b-it & input+output & safe/unsafe + rationale & NVIDIA OML + Gemma Terms \\
ShieldGemma~\cite{zeng2024shieldgemma} & 9B & Gemma2-9B & input+output & safe/unsafe & Gemma Terms of Use \\
GuardReasoner~\cite{liu2025guardreasoner} & 8B & Llama-3.1-8B & input+output+refusal & safe/unsafe + chain & Apache-2.0 \\
PolyGuard~\cite{kumar2025polyguard} & 7B & Qwen2.5-7B-Instruct & input+output+refusal & safe/unsafe & CC-BY-4.0 \\
Qwen3Guard~\cite{zhao2025qwen3guard} & 8B & Qwen3-8B & input+output+refusal & safe/unsafe/controversial & Apache-2.0 \\
GPT-OSS-SafeGuard~\cite{openai2025gptosssafeguard} & 20B & gpt-oss-20b & custom & custom & Apache-2.0 \\
YuFeng-XGuard~\cite{lin2026yufeng} & 8B & Qwen3-8B & input+output & safe/unsafe (threshold) & Apache-2.0 \\
\bottomrule
\end{tabular}}
\caption{Guardrails used in this paper.}
\label{tab:models}
\end{table*}

\section{Behavior-Side Ablations}
\label{app:behavior-ablation}

\subsection{Haystack Type: Priming Effect}

\begin{figure*}[h!]
  \centering
  \includegraphics[width=0.9\textwidth]{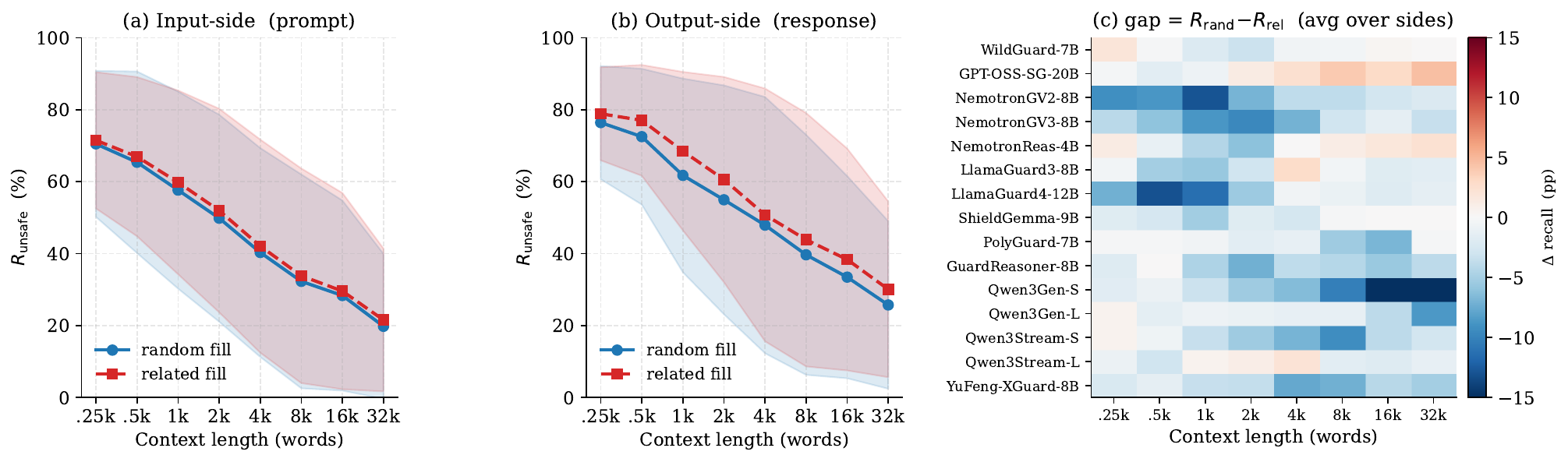}
  \caption{Effect of haystack type on $\Ru$. (a--b) Cross-model average $\Ru$ under prompt/response, Random vs.\ Related. (c) Per-model heatmap of $\mathrm{gap}=\Ru^{\mathrm{Random}}-\Ru^{\mathrm{Related}}$, averaged over the two audit sides; negative (blue) means Related $>$ Random.}
  \label{fig:relevance-gap}
\end{figure*}

Contrary to intuition, Figure~\ref{fig:relevance-gap} (a--b) shows that Related haystacks yield consistently higher $\Ru$ than Random across all eight lengths: the cross-model gap is between $-$1.73\% and $-$4.45\% (largest at 1k--2k), and the effect on response is 2--3$\times$ that on prompt. Figure~\ref{fig:relevance-gap} (c) shows that the effect is stable across the 15 models with only one positive outlier (GPT-OSS-SafeGuard-20B). We interpret it as a \emph{priming effect}: a haystack of the same topic as the needle keeps the ``risk-related'' semantic backdrop active and sustains a higher safety activation, whereas a Random haystack injects the needle into an unrelated semantic flow and lets attention dilute more easily.

\subsection{Needle-Position Sensitivity: Lost-in-the-Middle and Endpoint Asymmetry}

\begin{figure*}[h!]
  \centering
  \includegraphics[width=0.9\textwidth]{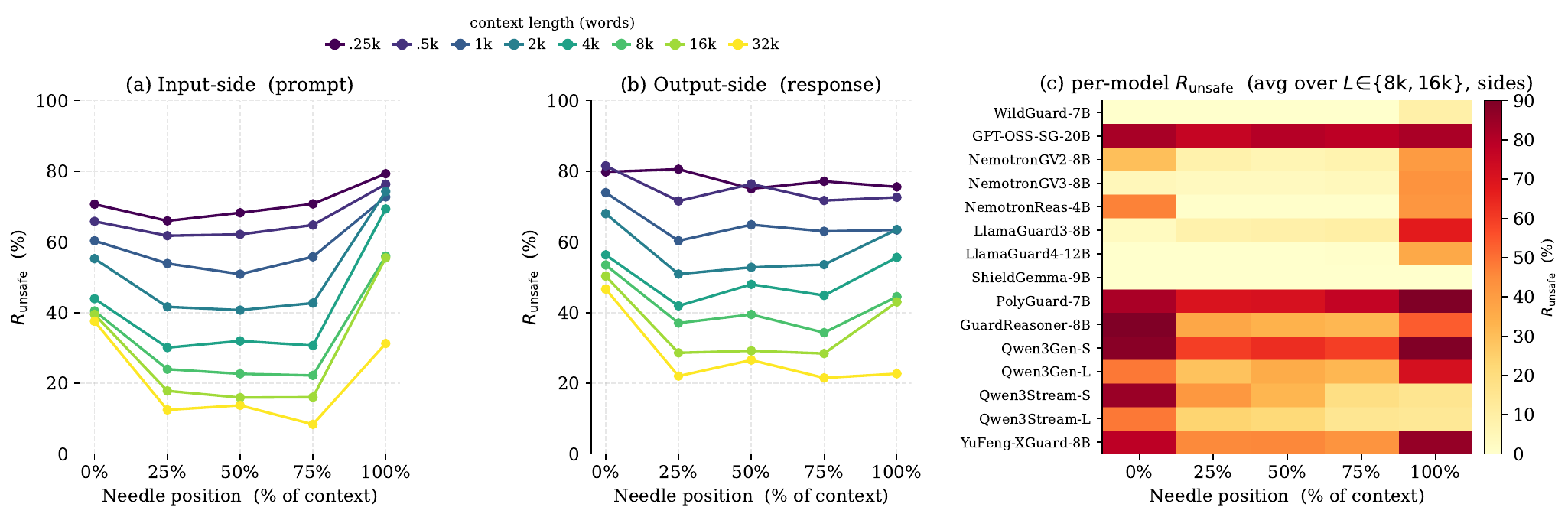}
  \caption{Sensitivity to needle position. (a--b) $\Ru$ as a function of needle position under prompt/response; one curve per length. (c) Per-model heatmap of $\Ru$ at $L\!\in\!\{8\text{k}, 16\text{k}\}$, averaged over audit side and haystack type.}
  \label{fig:position-recall}
\end{figure*}

As shown in Figure~\ref{fig:position-recall} (a--b), $\Ru$ at $p\!\in\!\{0\%, 100\%\}$ is higher than at the three middle positions across all lengths, exhibiting a typical lost-in-the-middle pattern. Taking the edge--middle gap as ($p$-0\%, $p$-100\% average) minus ($p$-25\%, $p$-50\%, $p$-75\% average), the gap grows monotonically from $+$6.67\% at 0.25k to $+$30.88\% at 16k on the prompt side, and from $+$0.10\% to $+$17.92\% on the response side. At 32k, the prompt side shows an extra collapse at $p\!=\!100\%$ (55.49 $\to$ 31.27) and the cross-model edge--middle gap falls from 24.40\% at 16k to 17.09\% at 32k. To avoid the mixed regime where endpoint advantage fails together with parsing failure, Figure~\ref{fig:position-recall} (c) restricts the slice to $\{8\text{k}, 16\text{k}\}$. The endpoint pattern is asymmetric: the prompt side is roughly bimodal, while the response side prefers the head -- $p\!=\!0\%$ is the highest at all lengths and $p\!=\!100\%$ falls to the middle. This matches the chat-template structure (prompt detection anchors both ends of a user turn to localize instructions; response detection mainly anchors the beginning of the assistant reply for safety cues). Figure~\ref{fig:position-recall} (c) further splits the models into three groups: (i) lost-in-the-middle (NemotronReasoning-4B 44.06\%, GuardReasoner-8B 38.04\%, YuFeng-XGuard-8B 37.58\%); (ii) long-context robust (GPT-OSS-SafeGuard-20B, PolyGuard-7B); (iii) total collapse (WildGuard-7B, ShieldGemma-9B), where all positions are near zero at 8k--16k.

\subsection{Failure Modes and Archetypes}

\begin{figure*}[h!]
  \centering
  \includegraphics[width=0.9\textwidth]{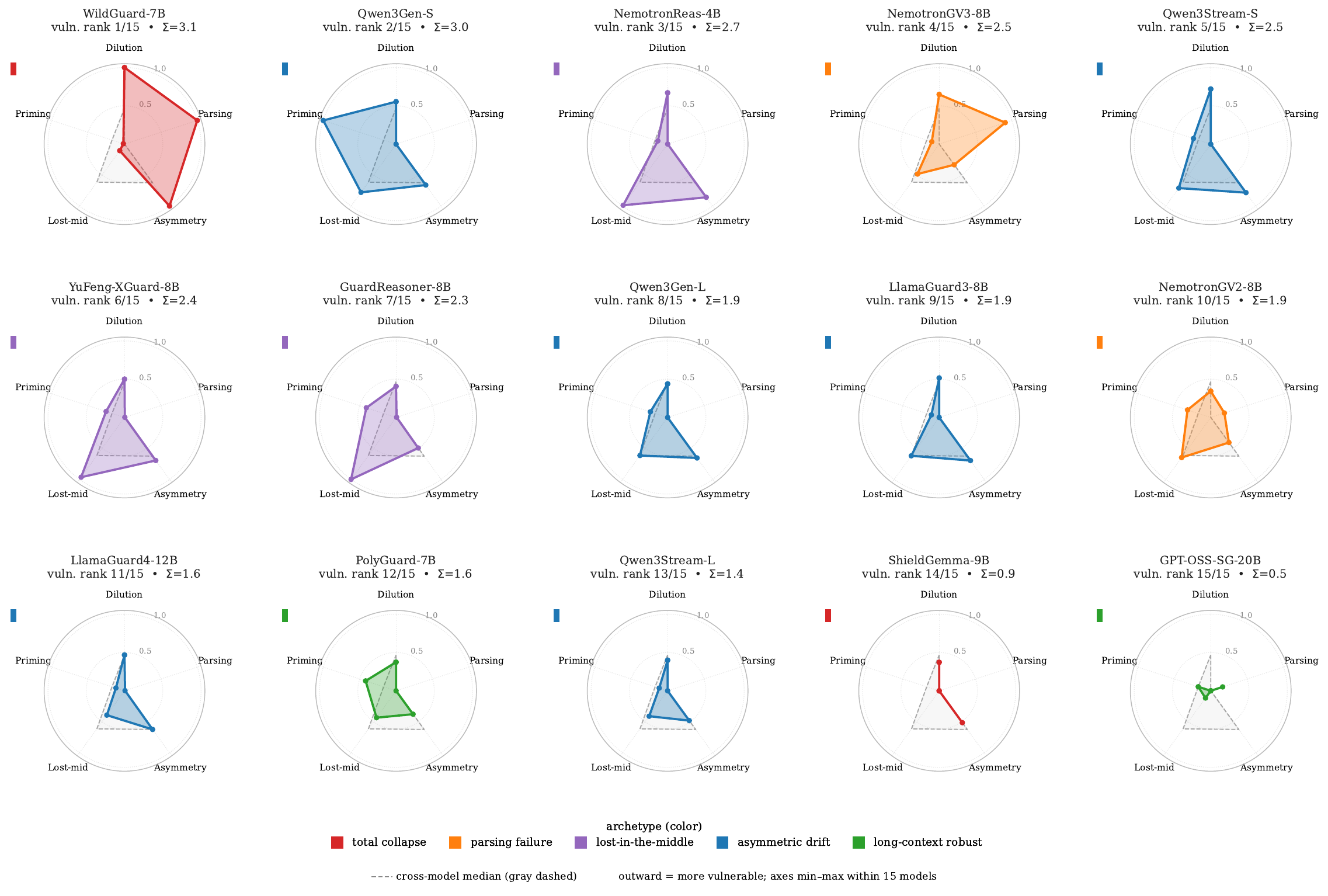}
  \caption{Radar of 15 guardrails over five failure axes. All axes are min--max normalized within the 15 models to $[0,1]$ (outside $=$ more brittle on that axis); subplots are sorted by the normalized total score, colored by archetype.}
  \label{fig:archetype}
\end{figure*}

To consolidate the observations at the model level, Figure~\ref{fig:archetype} merges five failure axes into a single radar: \emph{Dilution} ($\Delta\Ru$ on Benign-Fill), \emph{Parsing} (\texttt{inv@32k}), \emph{Asymmetry} (absolute difference of prompt/response $\Delta\Ru$), \emph{Lost-mid} (edge--middle gap at 8k--16k), and \emph{Priming} (Random$-$Related). The shapes split the 15 guardrails into five archetypes: (i) \textbf{total collapse} (WildGuard, ShieldGemma); (ii) \textbf{parsing failure} (NemotronGuardV3, WildGuard); (iii) \textbf{lost-in-the-middle} (NemotronReasoning, GuardReasoner, YuFeng-XGuard); (iv) \textbf{asymmetric drift} (Qwen3Guard-Stream, LlamaGuard4); (v) \textbf{long-context robust} (GPT-OSS-SafeGuard, PolyGuard), which serves as a natural upper-bound reference for our methods.

\section{Mediation Consistency Across the Three Layers}
\label{app:mediation}

We further check that the three layers move together beyond a common length covariate. On the 48 (model, length) units of Benign-Fill / unsafe, we pair the layer-level scalars (sample-mean $\Aneedle$, sample-mean logit margin, $\Ru$) and compute Pearson partial correlations after controlling for $\log_2 L$. The evidence is \emph{mediation consistency}, not strict causality: partial correlation rules out length as the most salient confounder but not other unobserved variables.

\begin{figure*}[h!]
  \centering
  \includegraphics[width=0.9\textwidth]{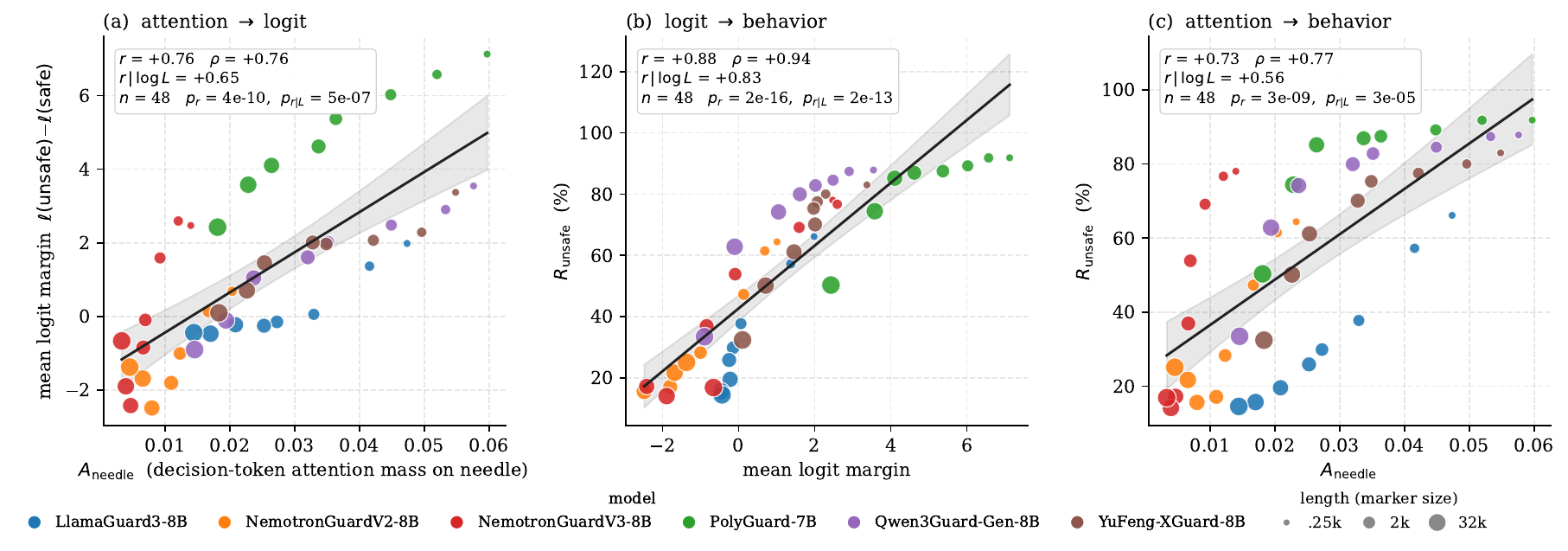}
  \caption{Correlations among attention, logit, and behavior on 48 (model, length) units. Each point is a unit; color encodes the model, and marker size scales with $\log_2 L$. Each panel reports Pearson $r$, Spearman $\rho$, and the partial correlation $r\!\mid\!\log L$.}
  \label{fig:corr-chain}
\end{figure*}

Figure~\ref{fig:corr-chain} shows that both legs of the chain are highly significant: attention$\to$logit $r=+0.76$ ($p=4.0\!\times\!10^{-10}$); logit$\to$behavior $r=+0.88$ ($p\!\approx\!2.0\!\times\!10^{-16}$) with a near-sigmoidal shape; end-to-end attention$\to$behavior $r=+0.73$. Partial correlations after controlling for $\log_2 L$ are $+0.65$, $+0.83$, and $+0.56$ (all $p<10^{-4}$), so the three layers continue to move together on the residual structure of the (model, length) grid.

\section{Per-Head Needle-Attention Heatmap: Full Length Sweep}
\label{app:headmap}

\begin{figure*}[t!]
  \centering
  \includegraphics[width=1.0\textwidth]{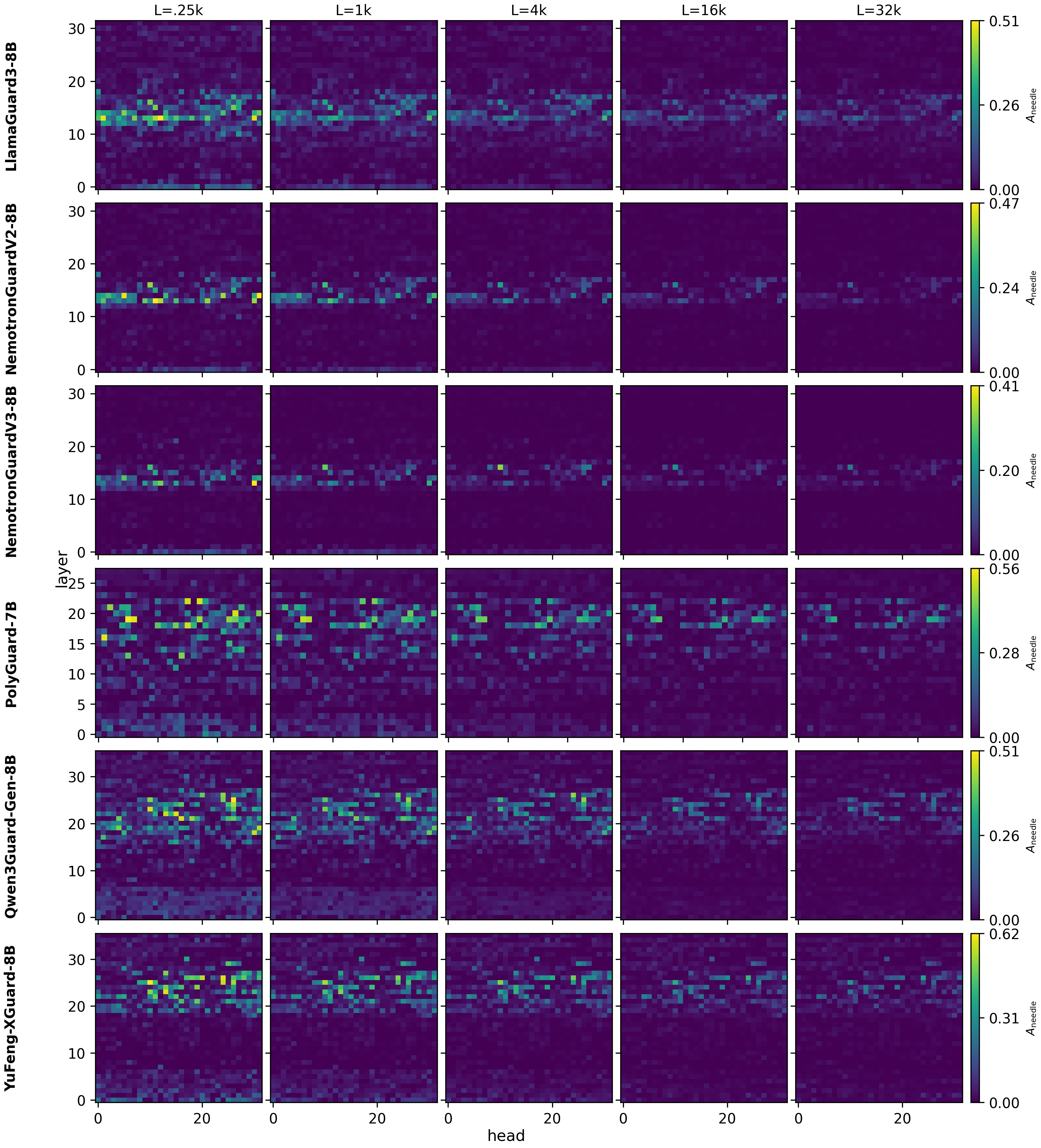}
  \caption{Per-head needle-attention heatmap, full length sweep (6 models $\times$ 5 lengths). Each row is one model; columns are $L\!\in\!\{0.25\text{k}, 1\text{k}, 4\text{k}, 16\text{k}, 32\text{k}\}$. Horizontal axis: head index; vertical axis: layer index; color: $\Aneedle^{(\ell,h)}$ (fill $=$ Benign with Random and Related pooled, label $=$ unsafe, prompt$+$response sample-level pooled). Each row shares its vmax and colorbar, so colors are directly comparable across lengths.}
  \label{fig:head-heatmap-full}
\end{figure*}

Figure~\ref{fig:head-heatmap-full} expands Figure~\ref{fig:head-heatmap} to all five lengths. Three patterns consistent with the main text are visible: (i) bright pixels remain sparsely concentrated on the same middle-late layers across all lengths, so the retrieval heads are structurally stable; (ii) the band darkens monotonically from 0.25k to 32k with no rebound; (iii) the band width differences across models correspond exactly to the vertical separation in the top-5\% cumulative curves of Figure~\ref{fig:head-topk}.

\begin{table*}[h!]
\centering
\resizebox{0.9\textwidth}{!}{%
\begin{tabular}{lcccccccc}
\toprule
Guard model & C1 top-5\%@256 (b/g/$\Delta$) & C1 top-5\%@32k (b/g/$\Delta$) & C2 $\rho$@32k & C2 Jaccard@32k & C3 $|\beta|$ ratio g/b & C3 MWU $p$ & Verdict \\
\midrule
LlamaGuard3-8B    & 0.288 / 0.335 / $+$4.7   & 0.200 / 0.337 / $\mathbf{+13.7}$ & 0.746 & 0.130 & $\mathbf{2.49}$ & $<10^{-15}$ & Partial \\
NemotronGuardV2-8B & 0.432 / 0.513 / $\mathbf{+8.1}$  & 0.306 / 0.445 / $\mathbf{+14.0}$ & 0.960 & 0.465 & 1.32 & $6\!\times\!10^{-4}$ & Partial \\
NemotronGuardV3-8B & 0.440 / 0.524 / $\mathbf{+8.4}$  & 0.313 / 0.512 / $\mathbf{+19.9}$ & 0.875 & 0.465 & 0.74 & 0.98 & Partial \\
PolyGuard-7B       & 0.307 / 0.330 / $+$2.3   & 0.279 / 0.429 / $\mathbf{+15.0}$ & 0.875 & 0.356 & 1.48 & $<10^{-15}$ & Partial \\
Qwen3Guard-Gen-8B  & 0.286 / 0.283 / $-$0.3   & 0.309 / 0.388 / $+$7.9          & 0.821 & 0.398 & $\mathbf{2.03}$ & $<10^{-15}$ & Partial \\
YuFeng-XGuard-8B   & 0.316 / 0.366 / $+$5.0   & 0.259 / 0.389 / $\mathbf{+13.1}$ & 0.787 & 0.094 & $\mathbf{2.19}$ & $<10^{-15}$ & Partial \\
\bottomrule
\end{tabular}}
\caption{Base vs.\ guard head-specificity comparison. Bold values reach the strong-specificity threshold (C1 $\Delta\!\geq\!5\%$; C3 $|\beta|$ ratio $>\!1.5$ with $p<0.05$).}
\label{tab:base-vs-guard}
\end{table*}

\section{Base vs.\ Guard Control Experiment}
\label{app:base-vs-guard}

The sparsity of $\Hsafety$ in Section~\ref{sec:retrieval-heads} could reflect a generic long-tail property of transformer attention (attention sinks, induction heads) rather than something specific to guard fine-tuning. This appendix runs a base--guard differential control to reject this null.

\paragraph{Setup.} For each guard in Section~\ref{sec:mechanism} we read its base from Appendix~\ref{app:models}; the three Llama guards share \texttt{Llama-3.1-8B-Instruct} (NemotronGuardV2 is in fact a LoRA adapter on this base), and PolyGuard/YuFeng-XGuard reuse the YaRN \texttt{rope\_scaling} (factor $=4$) so that bases do not collapse from naive RoPE extrapolation. Inputs (the Section~\ref{sec:degradation} unsafe-needle samples; Benign-Fill, 8 lengths, prompt and response) and attention extraction (Section~\ref{sec:mechanism}, $\qstar$ aligned with the guard side) are identical for base and guard; no logit/behavior is compared since the base lacks safe/unsafe surface forms. We measure three differentials, with outcomes (strong/partial / no specificity) pre-registered:

\begin{figure}[h!]
  \centering
  \includegraphics[width=\columnwidth]{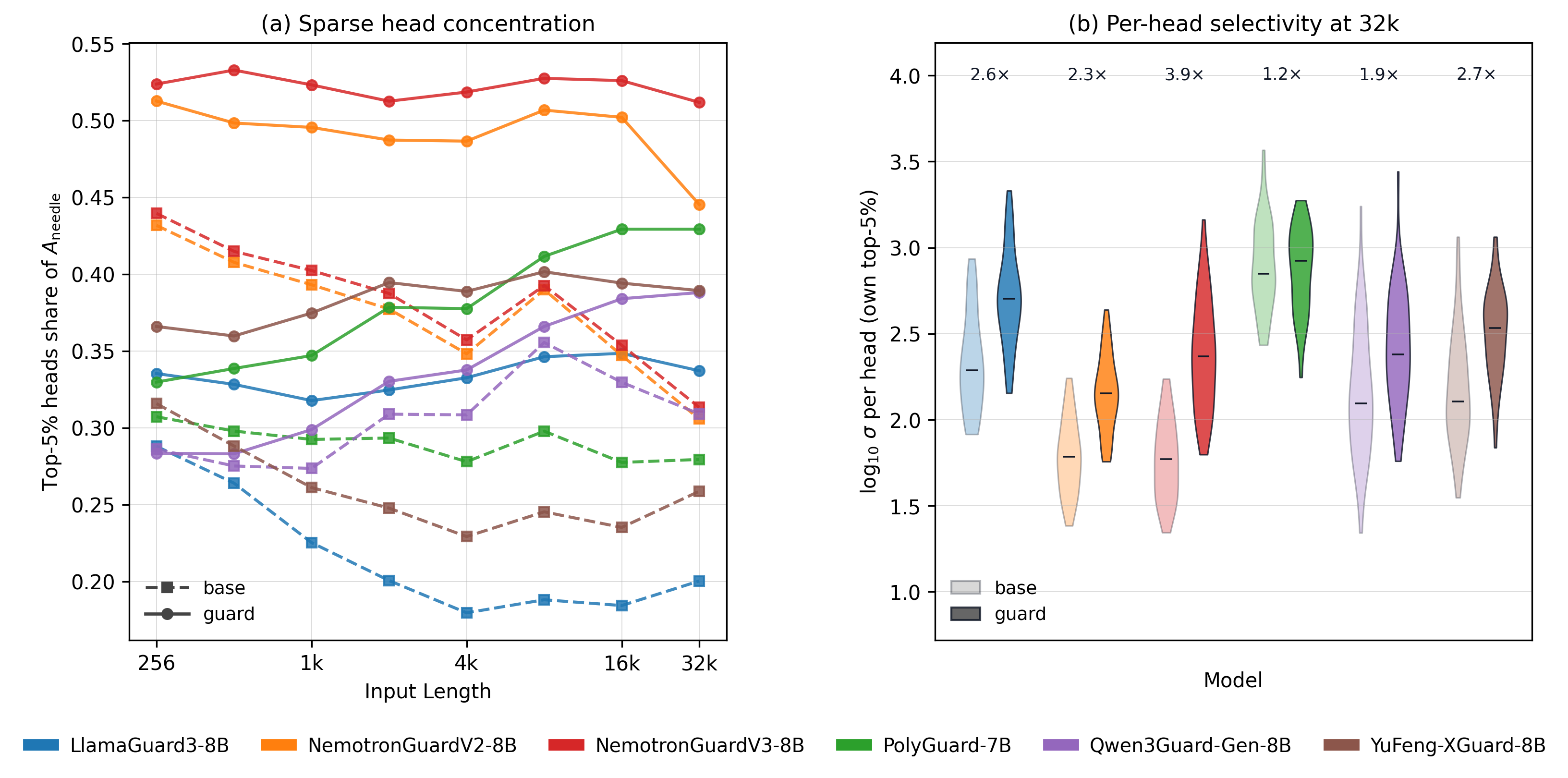}
  \caption{Shaping effect of guard fine-tuning on retrieval heads. (a) Top-5\% head share of $\Aneedle$ per (model, length); solid: guard, dashed: base. (b) Paired violin of $\log_{10}\sigma$ on top-5\% heads at $L\!=\!32k$.}
  \label{fig:head-evidence}
\end{figure}

\begin{itemize}
\item \textbf{C1: head concentration.} Share of needle attention taken by top-$k$ heads ($k\!\in\!\{1\%, 5\%, 10\%, 20\%\}$) per (model, length).
\item \textbf{C2: ranking consistency.} Spearman rank correlation $\rho_{\mathrm{rank}}$ between guard and base head rankings of $\Aneedle$ at $L\!=\!32k$; $\rho\!\approx\!1$ means the guard has not moved attention to different heads.
\item \textbf{C3: per-head dilution slope.} Regressing $\Aneedle^{(l,h)}$ on $\log_2 L$ over 256$\to$32k gives slope $\beta^{(l,h)}$; we run a Mann--Whitney test between base and guard.
\end{itemize}

\paragraph{Result: partial specificity (6/6).} All six pairs satisfy C2 $\rho$@32k $\geq 0.74$ (head identity highly overlaps, far above the strong threshold $\rho<0.7$), yet the guard top-5\% share at 32k is systematically $+$7.9\% to $+$19.9\% above the base, and C3 is significant on 5/6 pairs (Table~\ref{tab:base-vs-guard}, Figure~\ref{fig:head-evidence} (a)). When base and guard each use \emph{their own} top-5\% retrieval heads (a fairer absolute measurement), the guard is strictly larger on every pair, with cross-model$\times$audit-side ratios of 2.12$\times$--10.30$\times$ (Table~\ref{tab:base-vs-guard-abs}, Figure~\ref{fig:head-evidence} (b)). Guard fine-tuning therefore does not reassign attention to different heads but amplifies the magnitude on existing retrieval-bearing heads and steepens their long-$L$ dilution slopes -- the basis for restricting AHS's sharpening target (Section~\ref{sec:ahs}) to $\Hsafety$.

\begin{table}[t]
\centering
\resizebox{\columnwidth}{!}{%
\begin{tabular}{lccc}
\toprule
Guard model & base@32k & guard@32k & guard/base@32k \\
\midrule
LlamaGuard3-8B & 0.0178 & 0.0938 & 5.28$\times$ \\
NemotronGuardV2-8B & 0.0173 & 0.0390 & 2.26$\times$ \\
NemotronGuardV3-8B & 0.0175 & 0.0323 & 1.84$\times$ \\
PolyGuard-7B & 0.0571 & 0.1469 & 2.57$\times$ \\
Qwen3Guard-Gen-8B & 0.0362 & 0.1092 & 3.02$\times$ \\
YuFeng-XGuard-8B & 0.0314 & 0.1361 & 4.33$\times$ \\
\bottomrule
\end{tabular}}
\caption{Absolute mean $\Aneedle$ at 32k on each side's own top-5\% retrieval heads (sample-weighted merge over prompt and response).}
\label{tab:base-vs-guard-abs}
\end{table}

\section{Token-Level and Head-Level Sufficiency Interventions}
\label{app:intervention}

This appendix gives the full design of the two interventions summarized in Section~\ref{sec:intervention}. Both run on Benign-Fill unsafe samples at lengths 8k/16k/32k and aggregate by sample count.

\paragraph{Attention masking.} Three visibility conditions are compared on the same inputs: \texttt{full} (the original long context); \texttt{needle-only} (only the unsafe needle plus template tokens); \texttt{random} (an equal token budget filled with random haystack instead of the needle).

\paragraph{Retrieval-head amplification.} Only $\Hsafety$ is modified. We post-hoc amplify the softmax-output attention from $\qstar$ to the unsafe-needle span on the selected heads and renormalize:
\begin{equation}
\tilde A^{(\ell,h)}_{\qstar,j} = \frac{\bigl(1+(\alpha-1)\,\mathbf{1}[j\!\in\!\mathrm{needle}]\bigr)\,A^{(\ell,h)}_{\qstar,j}}{Z^{(\ell,h)}},
\label{eq:retrieval-amplify}
\end{equation}
where $(\ell,h)\!\in\!\Hsafety$, $\alpha\!\in\!\{2, 4, 8\}$, and $Z^{(\ell,h)}$ is the renormalization constant. We fix $K\!=\!5\%$ and use an equal number of random heads (\texttt{random-K}) as the control.

\begin{figure}[t]
  \centering
  \includegraphics[width=\columnwidth]{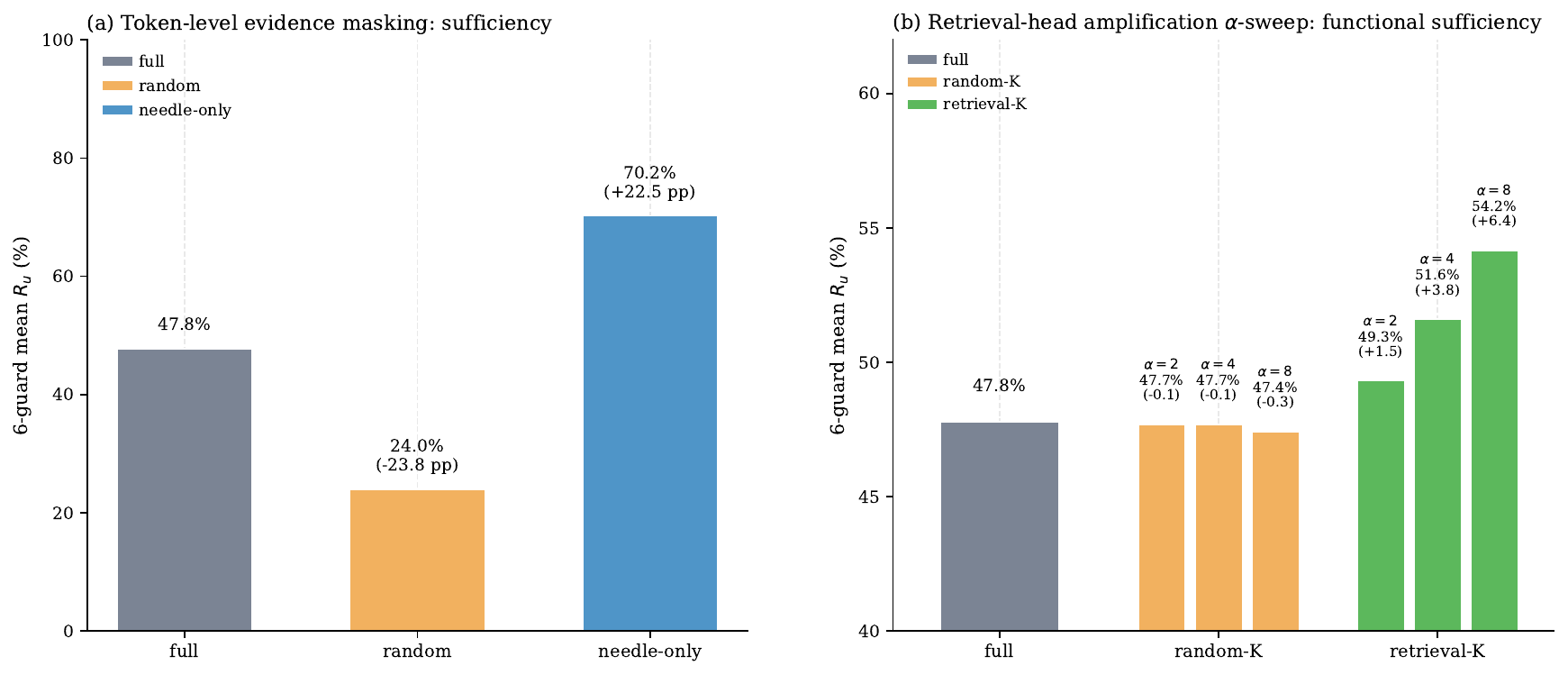}
  \caption{Sufficiency interventions (six-guard weighted, $L\!\in\!\{8\text{k}, 16\text{k}, 32\text{k}\}$, $K\!=\!5\%$). (a) Attention masking: \texttt{full} vs.\ \texttt{random} vs.\ \texttt{needle-only}. (b) Retrieval-head amplification at $\alpha\!\in\!\{2,4,8\}$ vs.\ \texttt{random-K}.}
  \label{fig:intervention}
\end{figure}

Figure~\ref{fig:intervention} (a): relative to the 47.8\% \texttt{full} baseline, \texttt{needle-only} climbs to 70.2\% ($+$22.5\%) while \texttt{random} falls to 24.0\% ($-$23.8\%). The recovery comes from preserving unsafe evidence rather than from shortening the input, so unsafe evidence itself is a sufficient signal for long-context safety judgment. Figure~\ref{fig:intervention} (b): \texttt{retrieval-K} monotonically improves with $\alpha$ ($+$1.5\%, $+$3.8\%, $+$6.4\% at $\alpha\!=\!2,4,8$) on all six guardrails, while the matched \texttt{random-K} stays within $[-0.3\%, -0.1\%]$. $\Hsafety$ provides a functional contribution to unsafe-evidence reading; arbitrary attention perturbations do not produce comparable gains.

\begin{table*}[t!]
\centering
\resizebox{\textwidth}{!}{%
\begin{tabular}{lccccccccc}
\toprule
Variant & use length & use type & use cascade & SafetyNIAH & MSJ & NINJA & Qwen3G-long & RShield-long & 5-bench mean \\
\midrule
S0 \texttt{original} & -- & -- & -- & 64.86 & 73.59 & 65.00 & 51.82 & 53.58 & 61.77 \\
S1 \texttt{fixed\_safety} (single best) & -- & -- & -- & 80.37 & 99.95 & 80.42 & 78.15 & 71.90 & 82.16 \\
S2 \texttt{+length} (length-only) & \checkmark & -- & -- & 81.15 & 100.00 & 80.21 & 78.44 & 73.34 & 82.63 \\
S3 \texttt{+length +cascade} & \checkmark & -- & \checkmark & 82.04 & 100.00 & 80.82 & 78.39 & 73.34 & 82.92 \\
S4 \texttt{+length +type} & \checkmark & \checkmark & -- & 81.79 & 100.00 & 80.85 & 78.49 & 76.06 & 83.44 \\
\rowcolor{gray!15}
\textbf{S5 \texttt{+length +type +cascade} $=$ CAHR} & \checkmark & \checkmark & \checkmark & \textbf{82.03} & \textbf{100.00} & \textbf{81.25} & \textbf{78.47} & \textbf{76.03} & \textbf{83.56} \\
\bottomrule
\end{tabular}}
\caption{Component ablation of CAHR on the CD family; the AHS family gives the same conclusion (CAHR 5-bench mean $=$ 74.33).}
\label{tab:cahr-ablation}
\end{table*}

\section{Hyperparameter Sweep and CAHR Ablation}
\label{app:cahr-sweep}

\subsection{Length-Wise Sweep on SafetyNIAH}

Tables~\ref{tab:sweep-cd} and \ref{tab:sweep-ahs} give the $\Funsafe$ of 17 candidate configurations (base $+$ CD $\times 4$ $+$ AHS $\times 12$) on the 8 length buckets of SafetyNIAH.

\begin{table}[h!]
\centering
\resizebox{\columnwidth}{!}{%
\begin{tabular}{lcccccccc}
\toprule
Candidate & 0.25k & 0.5k & 1k & 2k & 4k & 8k & 16k & 32k \\
\midrule
orig ($\varnothing$) & \textbf{82.71} & 80.67 & 73.75 & 66.74 & 59.76 & 52.68 & 49.16 & 40.21 \\
CD-64 & 81.47 & 81.01 & 79.86 & 79.84 & 79.53 & 78.45 & 77.28 & 76.06 \\
CD-128 & 81.92 & \textbf{80.93} & \textbf{79.90} & \textbf{79.95} & \textbf{79.77} & \textbf{79.04} & \textbf{78.38} & 77.36 \\
CD-256 & 81.70 & 80.93 & 79.20 & 79.21 & 79.37 & 78.82 & 78.27 & \textbf{78.10} \\
CD-512 & 81.70 & 79.95 & 77.78 & 78.03 & 78.24 & 77.78 & 77.71 & 77.54 \\
\bottomrule
\end{tabular}}
\caption{CD sweep (six-guard mean $\Funsafe$); per-column bold is the CD-family argmax. The orig row is bold as the $\varnothing$ reference.}
\label{tab:sweep-cd}
\end{table}

\begin{table}[h!]
\centering
\resizebox{\columnwidth}{!}{%
\begin{tabular}{lcccccccc}
\toprule
$(K,\tau)$ & 0.25k & 0.5k & 1k & 2k & 4k & 8k & 16k & 32k \\
\midrule
(1\%, 0.1) & 83.86 & 82.00 & 76.97 & 72.18 & 68.60 & 61.61 & 56.50 & 48.86 \\
(1\%, 0.3) & 84.07 & 82.31 & 77.29 & 72.44 & 68.57 & 61.83 & 56.74 & 49.85 \\
(1\%, 0.5) & 84.00 & 82.21 & 77.14 & 71.96 & 67.88 & 61.20 & 56.37 & 49.60 \\
(1\%, 0.7) & 83.64 & 81.83 & 76.76 & 71.11 & 66.75 & 59.76 & 54.98 & 48.10 \\
(5\%, 0.1) & 82.25 & 80.36 & 78.10 & 73.92 & 71.40 & 66.39 & 62.29 & 50.78 \\
(5\%, 0.3) & 83.83 & 82.51 & 79.20 & 75.48 & 71.81 & 66.85 & 63.62 & 52.41 \\
(5\%, 0.5) & \textbf{84.32} & 83.03 & 79.57 & 75.36 & 71.86 & 66.16 & 62.00 & 51.45 \\
(5\%, 0.7) & 84.12 & 82.64 & 78.64 & 73.91 & 70.47 & 63.83 & 58.95 & 49.83 \\
(10\%, 0.1) & 79.04 & 77.16 & 73.99 & 70.68 & 65.84 & 62.66 & 60.51 & 48.60 \\
(10\%, 0.3) & 82.11 & 81.29 & 78.93 & 74.29 & 70.57 & 65.38 & 65.11 & 54.45 \\
(10\%, 0.5) & 83.67 & \textbf{83.06} & \textbf{80.39} & \textbf{76.84} & \textbf{74.20} & \textbf{67.73} & \textbf{65.36} & \textbf{56.22} \\
(10\%, 0.7) & 84.08 & 82.97 & 79.60 & 75.23 & 72.30 & 65.24 & 61.21 & 52.65 \\
\bottomrule
\end{tabular}}
\caption{AHS sweep (six-guard mean $\Funsafe$); per-column bold is the AHS-family argmax.}
\label{tab:sweep-ahs}
\end{table}

\emph{No single configuration is universally optimal.} In the CD family the argmax is $W\!=\!128$ on $L\!\in\![0.5\text{k}, 16\text{k}]$ and switches to $W\!=\!256$ at $L\!=\!32\text{k}$. In the AHS family the argmax shifts from $(K\!=\!5\%, \tau\!=\!0.5)$ at $L\!=\!0.25\text{k}$ to $(K\!=\!10\%, \tau\!=\!0.5)$ at $L\!\geq\!0.5\text{k}$. At $L\!=\!0.25\text{k}$, orig even outperforms all four CD configurations (82.71 vs.\ $\leq 81.92$). The length-conditional CAHR routing table (with $\varnothing$ included as a candidate) is designed to absorb such drift. We only show the six-guard mean here; per-guard drift is larger.

\subsection{Component Ablation of CAHR}
\label{app:cahr-ablation}

We split CAHR into six independently switchable routing strategies, from the most naive \texttt{original} (no method) to the full \texttt{CAHR} (length$\times$type table $+$ base-cascade) layered one at a time. All routing tables are fitted on SafetyNIAH only; the five evaluation benchmarks (SafetyNIAH / MSJ / NINJA / Qwen3G-long / RShield-long) are frozen across variants. All numbers are six-guard averages; SafetyNIAH / Qwen3G-long / RShield-long report $\Funsafe$, and MSJ / NINJA report $\Ru$. The last column is the 5-benchmark mean.

\paragraph{Reading by component.}
\begin{itemize}
\item \textbf{S0$\to$S1 ($+$20.4\%).} A single best configuration on SafetyNIAH already brings most of the gain; this is the zero-order benefit of any long-context guardrail method.
\item \textbf{S1$\to$S2 ($+$0.5\%).} Replacing the single global best with per-length-bucket bests adds $0.3\%$--$1.5\%$ on long-context dual-class benchmarks (Qwen3G-long, RShield-long), confirming the cross-length drift.
\item \textbf{S2$\to$S3 ($+$0.3\%).} Base-cascade rescues samples whose base \texttt{unsafe} decision would be flipped by the routed method on attack-only benchmarks (NINJA $+$0.6\%).
\item \textbf{S2$\to$S4 ($+$0.8\%).} Adding the type dimension lifts the response-heavy RShield-long by $+$2.7\% (the largest single-item gain).
\item \textbf{S4$\to$S5 $=$ CAHR ($+$0.1\%).} Cascade and type stacking take the last small bites of NINJA and SafetyNIAH.
\end{itemize}

\begin{figure*}[t!]
  \centering
  \includegraphics[width=1.0\textwidth]{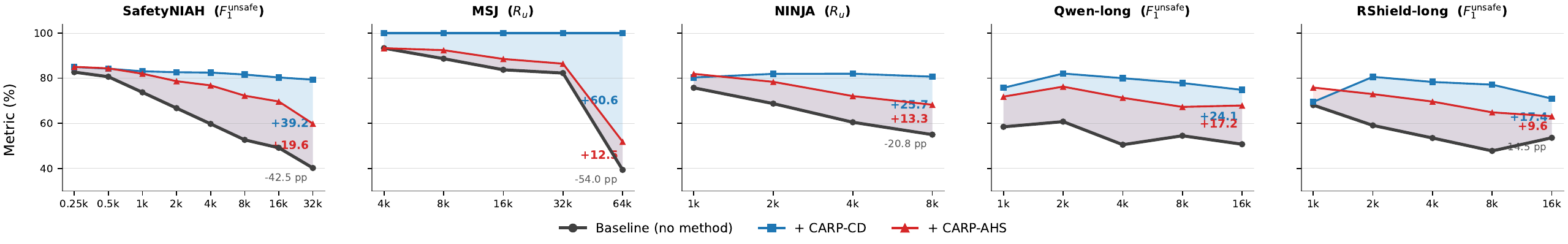}
  \caption{Length-wise gain of CAHR on the five primary benchmarks (six-guard mean). The three curves are base, CAHR-CD, and CAHR-AHS; the horizontal axis is context length (words, log), and the vertical axis is the primary metric. Annotations in the longest bucket show the base drop and the CAHR recovery relative to the same-length base.}
  \label{fig:length-gain}
\end{figure*}

\begin{table*}[h!]
\centering
\resizebox{\textwidth}{!}{%
\begin{tabular}{llcccc}
\toprule
Benchmark & Method & $\Funsafe$ & Unsafe precision & Safe recall & FPR \\
\midrule
\multirow{3}{*}{SafetyNIAH (test)}
  & base     & 64.86 $[64.2, 65.7]$ & 82.39 $[81.5, 83.3]$ & 83.08 $[82.4, 83.8]$ & 16.92 $[16.2, 17.7]$ \\
  & CAHR-CD  & 82.03 $[81.4, 82.6]$ & 80.31 $[79.4, 81.2]$ & 70.60 $[69.6, 71.5]$ & 29.40 $[28.5, 30.4]$ \\
  & CAHR-AHS & 76.41 $[75.8, 77.0]$ & 82.65 $[81.8, 83.5]$ & 78.06 $[77.2, 78.9]$ & 21.94 $[21.1, 22.8]$ \\
\midrule
\multirow{3}{*}{Qwen3G-long}
  & base     & 51.82 $[49.6, 53.8]$ & 66.74 $[64.8, 68.6]$ & 73.67 $[72.1, 75.1]$ & 26.33 $[24.9, 27.9]$ \\
  & CAHR-CD  & 78.47 $[76.4, 80.2]$ & 77.14 $[74.8, 79.4]$ & 68.47 $[66.3, 70.7]$ & 31.53 $[29.3, 33.7]$ \\
  & CAHR-AHS & 71.41 $[69.2, 73.4]$ & 80.61 $[78.6, 82.8]$ & 79.56 $[78.0, 81.2]$ & 20.44 $[18.9, 22.0]$ \\
\midrule
\multirow{3}{*}{RShield-long}
  & base     & 53.58 $[51.8, 55.2]$ & 62.71 $[61.0, 64.3]$ & 75.15 $[74.3, 76.0]$ & 24.85 $[24.0, 25.7]$ \\
  & CAHR-CD  & 76.03 $[74.3, 77.7]$ & 67.77 $[65.5, 70.1]$ & 76.27 $[74.9, 77.7]$ & 23.73 $[22.3, 25.1]$ \\
  & CAHR-AHS & 68.81 $[66.9, 70.4]$ & 70.81 $[68.5, 73.1]$ & 84.21 $[83.3, 85.3]$ & 15.79 $[14.7, 16.7]$ \\
\bottomrule
\end{tabular}}
\caption{Full panel on dual-class benchmarks (six-guard mean, \%). Each cell is the point estimate with its 95\% sample-level bootstrap CI in brackets. SafetyNIAH is the held-out test split.}
\label{tab:deployment-panel}
\end{table*}

\subsection{5-Fold Cross-Validation on SafetyNIAH}
\label{app:cahr-devtest}

Section~\ref{sec:eval} fits the CAHR routing table on the dev split of SafetyNIAH but evaluates it only on the held-out test split. Table~\ref{tab:cahr-devtest} corroborates this fixed split with an independent 5-fold cross-validation.

\begin{table}[h!]
\centering
\resizebox{\columnwidth}{!}{%
\begin{tabular}{lccc}
\toprule
Method & Dev & Test & 5-fold CV (mean$\pm$std) \\
\midrule
CAHR-CD & 82.37 & 82.03 & 82.19$\pm$0.38 \\
CAHR-AHS & 76.97 & 76.41 & 76.64$\pm$0.32 \\
\bottomrule
\end{tabular}}
\caption{CAHR on SafetyNIAH: fixed dev/test split vs.\ 5-fold cross-validation ($\Funsafe$, six-guard mean, \%).}
\label{tab:cahr-devtest}
\end{table}

The fixed-split test number and the 5-fold CV mean agree closely for both families (CD: 82.03 vs.\ 82.19$\pm$0.38; AHS: 76.41 vs.\ 76.64$\pm$0.32), well within one cross-validation standard deviation, so the Table~\ref{tab:main-method} numbers are stable across held-out partitions of SafetyNIAH.

\section{Length-wise Gain}
\label{app:length-gain}

Figure~\ref{fig:length-gain} resolves the gains of Table~\ref{tab:main-method} by context length. The base drops monotonically with length across all benchmarks; CAHR-CD and CAHR-AHS recover most of that drop, confirming that the methods primarily restore the long-context dimension.

We take 1k words as the short-context cutoff based on Table~\ref{tab:source-data} (the maximum average length of existing guardrail benchmarks). Figure~\ref{fig:length-gain} likewise shows that CAHR-CD and CAHR-AHS stay at or above the base guard up to 1k, so the methods do not degrade short-context performance.

\section{Deployment Metrics, Cost, and Robustness}
\label{app:deployment}

\paragraph{Full metric panel.} Table~\ref{tab:deployment-panel} reports $\Funsafe$, unsafe precision, safe recall, and FPR on the three dual-class benchmarks. $\Funsafe$ rises under both CAHR-CD and CAHR-AHS, while unsafe precision stays close to or above the base guard.

\paragraph{Cost by length.} Table~\ref{tab:deployment-cost} reports the relative processed-token budget ($\times$ a single original call) of the deployed CAHR policy on SafetyNIAH. Cost rises with length: CAHR-CD grows from $1.44\times$ at 0.25k to $2.94\times$ at 32k, while CAHR-AHS stays cheaper ($1.48\times$ to $1.76\times$). AHS is the cost-conservative option; CD costs more and yields the larger $\Funsafe$ gains of Table~\ref{tab:deployment-panel}.

\paragraph{CD false-alarm growth.} Table~\ref{tab:deployment-fpr} shows that CD FPR on SafetyNIAH rises monotonically with the chunk count $N=\lceil L/W\rceil$. This is a controlled trade-off rather than a net regression: CD trades some safe recall for a large $\Funsafe$ gain (Table~\ref{tab:deployment-panel}), while AHS keeps FPR closer to the base. CAHR's per-length window selection avoids the small-window / many-chunk settings that drive the worst FPR growth.

\begin{table}[t!]
\centering
\begin{tabular}{lcc}
\toprule
$L$ & CAHR-CD & CAHR-AHS \\
\midrule
0.25k & 1.44 & 1.48 \\
1k    & 1.97 & 1.55 \\
4k    & 2.86 & 1.65 \\
16k   & 2.88 & 1.74 \\
32k   & 2.94 & 1.76 \\
\bottomrule
\end{tabular}
\caption{Relative processed-token budget of CAHR on SafetyNIAH by context length ($\times$ a single original call).}
\label{tab:deployment-cost}
\end{table}

\begin{table}[h!]
\centering
\resizebox{\columnwidth}{!}{%
\begin{tabular}{lcccc}
\toprule
$L$ & $W{=}64$ & $W{=}128$ & $W{=}256$ & $W{=}512$ \\
\midrule
0.25k & 21.78 (4) & 17.77 (2) & 14.22 (1) & 14.22 (1) \\
1k    & 30.78 (16) & 24.94 (8) & 20.76 (4) & 16.61 (2) \\
4k    & 44.27 (64) & 35.15 (32) & 27.79 (16) & 22.15 (8) \\
16k   & 64.20 (256) & 48.88 (128) & 39.08 (64) & 29.72 (32) \\
32k   & 75.45 (512) & 57.72 (256) & 45.40 (128) & 34.93 (64) \\
\bottomrule
\end{tabular}}
\caption{CD FPR on SafetyNIAH by length (six-guard mean, \%); the chunk count $N=\lceil L/W\rceil$ is in parentheses.}
\label{tab:deployment-fpr}
\end{table}

\paragraph{When to use CD vs.\ AHS.} CD and AHS trade off differently. CD treats the guard as a black box, applies to most guardrails, and yields the larger $\Funsafe$ gain at higher token cost and higher FPR (Tables~\ref{tab:deployment-panel}--\ref{tab:deployment-fpr}). AHS requires white-box access to attention, applies only to open-weight guards, and is the cheaper, lower-FPR option ($\sim\!1.5\times$ tokens; Table~\ref{tab:deployment-cost}). Prefer AHS when the latency or token budget is tight; prefer CD when the guard is closed or when maximizing recall is the priority.

\section{Use of AI Assistants}
This work uses AI assistants in a limited capacity for language refinement and academic polishing. Their use is restricted to improving readability and presentation and does not influence the research methodology, data analysis, or scientific claims. The authors make all substantive contributions.

\end{document}